\documentclass{article}

\PassOptionsToPackage{numbers,sort&compress}{natbib}
\usepackage[main, final]{neurips_2026}

\makeatletter
\@ifpackageloaded{inputenc}{}{\usepackage[utf8]{inputenc}}
\@ifpackageloaded{fontenc}{}{\usepackage[T1]{fontenc}}
\@ifpackageloaded{hyperref}{}{\usepackage{hyperref}}
\@ifpackageloaded{xcolor}{}{\usepackage{xcolor}}
\makeatother
\usepackage{url}
\usepackage{booktabs}
\usepackage{amsfonts}
\usepackage{nicefrac}
\usepackage{microtype}
\usepackage{graphicx}
\usepackage{xspace}
\usepackage{wrapfig}
\usepackage{amsmath}
\usepackage{mathtools}
\usepackage{algorithm}
\usepackage{algpseudocode}
\usepackage{caption}
\usepackage{minitoc}
\usepackage{titletoc}
\usepackage{amssymb}
\usepackage{multirow}
\usepackage{placeins}

\definecolor{lcpurple}{RGB}{147,112,219}
\definecolor{hlalpha}{RGB}{0,170,200}
\definecolor{hlz}{RGB}{155,40,180}
\definecolor{hltau}{RGB}{30,170,90}

\floatstyle{ruled}
\restylefloat{algorithm}

\definecolor{lightpurple}{RGB}{150, 110, 220}
\definecolor{linkcyan}{HTML}{0B8FA6}
\definecolor{citepurple}{HTML}{7B4DFF}
\definecolor{hgreen}{HTML}{2CA02C}
\definecolor{cred}{HTML}{C44E52}
\definecolor{apurple}{HTML}{8172B2}
\definecolor{ocyan}{HTML}{17BECF}

\definecolor{figgreen}{HTML}{3BAA35}
\definecolor{figsalmon}{HTML}{D98A5B}
\definecolor{figred}{HTML}{D62728}
\definecolor{figpurple}{HTML}{8B6FA6}
\definecolor{figcyan}{HTML}{139E9A}

\definecolor{lclgreen}{HTML}{3BAA35}
\definecolor{lclsalmon}{HTML}{D98A5B}
\definecolor{lclred}{HTML}{D62728}
\definecolor{lclpurple}{HTML}{8B6FA6}
\definecolor{lclcyan}{HTML}{139E9A}

\newcommand{\hfig}[1]{\textcolor{figgreen}{$#1$}}
\newcommand{\sfig}[1]{\textcolor{figsalmon}{$#1$}}
\newcommand{\cfig}[1]{\textcolor{figred}{$#1$}}
\newcommand{\afig}[1]{\textcolor{figpurple}{$#1$}}

\definecolor{hlbeta}{HTML}{8E44AD}

\hypersetup{
    colorlinks=true,
    citecolor=lightpurple,
    linkcolor=cyan,
    urlcolor=cyan,
}

\title{Meta-Reinforcement Learning with Zero-Shot Reinforcement Learning}

\author{%
  Jake Grigsby
  \quad
  Siddhant Agarwal$^\dagger$
  \quad
  Yu Lei$^\dagger$
  \quad
  Leonidas Varveropoulos$^\dagger$
  \quad
  Yuke Zhu \\
  The University of Texas at Austin \\
  $^\dagger$Equal contribution \\
  \texttt{grigsby@cs.utexas.edu}
}

\begin{document}

\maketitle

\begin{abstract}
Meta-reinforcement learning (meta-RL) trains agents to adapt to novel tasks from test-time experience, but requires diverse training environments and reward functions. While environments are becoming cheaper to generate, it remains difficult to anticipate every objective an agent may be asked to pursue at test time. Behavior Foundation Models (BFMs) learn policies from reward-free data and support zero-shot RL (ZSRL): adaptation to new objectives at test time using a large reward-labeled dataset. Recent work has shown that BFM task inference can instead be performed online without prior task-specific data. We note that this capability positions BFMs and meta-RL as alternative approaches for online adaptation to new reward functions \textit{in a fixed environment}. We then ask whether we can extend BFMs to compete with meta-RL more broadly, including in novel environments, and identify four challenges: environment identification, online data collection, test-time exploration, and learning from mixed reward supervision. We study each individually, then propose \texttt{MetaBFM}, a hybrid method that addresses all four. During training, it combines supervised reward-following with unsupervised learning; at test time, it can interpolate between exploration, exploitation, and ZSRL-style reward inference. We demonstrate that \texttt{MetaBFM} can recover the in-distribution performance of supervised meta-RL while generalizing to reward functions outside the training set. Our work suggests that BFMs and meta-RL are complementary research directions whose combined strengths allow adaptive agents to learn more diverse behaviors from a fixed set of supervised objectives.
\end{abstract}
\section{Introduction}

Meta-reinforcement learning agents improve on novel tasks by adapting their decision-making from test-time experience \cite{beck2023survey}. This capability relies on diverse training sets of environments and reward functions, and benefits from online data collection over many trials. Despite much algorithmic progress, meta-RL remains limited by the cost of generating these training sets. Most work focuses on toy problems where similar environments and/or reward functions are easily sampled; large-scale applications like AdA \cite{team2023human}, where a team of developers repurposed a commercial game engine to generate complex tasks and curricula, are rare exceptions. Generating a diverse set of learnable \textit{reward functions} that covers the desired range of behavior may be inherently difficult. Reward functions are challenging to design at scale \cite{booth2023perils, knox2023reward}, and imperfect proxies can break down under optimization pressure \cite{karwowski2023goodhart}. However, with hardware-accelerated simulators \cite{makoviychuk2021isaac, mittal2025isaac}, LLM-assisted coding \cite{karten2026automatic}, and promptable world models \cite{bruce2024genie, che2024gamegen}, the cost of generating \textit{environments} is rapidly declining.

Behavior Foundation Models (BFMs) are an emerging class of unsupervised RL methods \cite{laskin2021urlb} that train in a reward-free environment by learning to maximize many possible objectives \cite{agarwal2025a}. At test time, a target objective is specified by model inference on a large dataset of trajectories labeled with a particular reward function. BFMs' unsupervised training and supervised testing datasets are loosely analogous to the pretraining and finetuning paradigm of foundation models in other areas \cite{zhou2025comprehensive}. This adaptation process is often called zero-shot RL (ZSRL) \cite{touati2022does} and has shown promising results in domains like humanoid locomotion \cite{tirinzonizero, li2025bfm}. Meta-RL and BFMs both enable generalization without additional training, but ZSRL requires a large supervised adaptation dataset in advance, while meta-RL assumes the policy actively collects its own (far smaller) dataset. This distinction makes the two approaches applicable to different types of adaptation problems, and their test-time performance is difficult to compare directly. However, recent work by \citet{rupf2025optimistic} demonstrated that ZSRL task inference can be done online, without a preexisting dataset, under the same conditions as meta-RL. As a result, meta-RL and BFMs can now be considered direct competitors on problems that require adaptation to novel reward functions \textit{in a fixed environment}.

If BFMs can match a broader range of meta-RL's capabilities, they could help extend generalist RL to domains where train-time reward supervision covers only a narrow range of behaviors. We argue that this requires BFMs to (1) identify and generalize to novel environments, (2) learn effectively without a pre-collected training dataset, (3) make exploration-exploitation tradeoffs at test time, and (4) benefit from available reward supervision. We study each of these subproblems in turn with detailed experiments on a diagnostic toy problem and simulated humanoid locomotion before arriving at a unified framework that addresses all four. The resulting method, \texttt{MetaBFM}, trains a policy that spans meta-RL, ZSRL, and intrinsic exploration. During training, \texttt{MetaBFM} combines supervised reward-following with unsupervised learning while collecting training data online. At test time, we are free to interpolate between exploration, exploitation, and ZSRL-style reward inference. This hybrid method draws on the strengths of meta-RL and ZSRL: it retains meta-RL's ability to adapt within the training distribution while opening the possibility of generalizing to out-of-distribution reward functions. Our work suggests that BFMs and meta-RL are complementary research directions and that, together, they can shift the emphasis from specifying additional training objectives to designing diverse environments.
\section{Background}

\subsection{Meta-Reinforcement Learning}
\label{sec:background:metarl}

A meta-RL policy $\pi$ can be evaluated by its cumulative return over the first $k$ attempts, or episodes, in a new task. This objective creates an exploration-exploitation tradeoff at test time: the policy may sacrifice performance in early attempts to learn more about the task if exploiting that knowledge later leads to higher cumulative return \cite{zintgraf2022fast}. Methods differ in whether this tradeoff is managed by the designer as a hyperparameter \cite{finn2017model} or learned by the policy itself \cite{duan2016rl}. Policies may also be given $k$ free episodes to explore without consequence before being evaluated on attempt $k+1$; this \textbf{$k$-shot} setting creates explicit exploration and exploitation phases \cite{liu2021decoupling, stadie2018some}.

Meta-RL's test-time adaptation is enabled by training on a diverse set of tasks. In this paper, a ``task'' refers specifically to a combination of an environment and an objective---or, equivalently, to a pair of parameters $(c_d, c_r)$ that alter the transition dynamics and reward function, respectively. The basic goal of a context-based meta-learner is to use the multi-episodic trajectory\footnote{The main text assumes common MDP RL notation. Appendix \ref{app:background:rl} provides a reference and a standard introduction.} of interaction with the current task so far, $\tau_{:t} = \{s_0, a_0, r_1, s_1, a_1, \dots, s_t\}$, to infer the unknown $(c_d, c_r)$, exploit the unique characteristics of the task, and maximize returns. Methods broadly differ in whether the parameters $(c_d, c_r)$ are estimated by supervised regression to ground-truth values \cite{humplik2019meta}, as the byproduct of a self-supervised objective such as dynamics modeling \cite{zintgraf2021varibad}, or implicitly through the objective of maximizing returns \cite{wang2016learning}. In each case, the idea is to establish a training objective that is minimized by correctly distinguishing $(c_d, c_r)$ and then conditioning $\pi$ on the resulting representation.

Meta-RL is at its best when we can afford sample-inefficient online data collection \cite{team2023human, nikulin2024xland, lu2023structured}, which may be acceptable when we are willing to trade train-time compute for efficient test-time adaptation. Meta-learning is harder when restricted to a fixed dataset \cite{pong2022offline, dorfman2020offline, gao2023context, li2024towards}, as it suffers from an extreme form of the distribution shift problem that defines offline RL more broadly \cite{levine2020offline}. Methods also differ in whether task rewards are available during meta-training. Standard meta-RL is \textit{supervised} by a distribution of reward-labeled tasks, whereas \textit{unsupervised} meta-RL learns from reward-free interaction and only encounters rewards at test time \cite{beck2023survey}. We focus on online meta-RL that spans both regimes, meta-training on a set of reward functions while simultaneously learning a reward-free BFM that supports ZSRL-style task inference at test time. This hybrid framing leads to a method distinct from prior unsupervised meta-RL \cite{gupta2018unsupervised, jabri2019unsupervised, pappalardo2026unsupervised}, which omits reward supervision entirely, and from prior ZSRL/BFM work, which does not meta-train across environments.

\subsection{Behavior Foundation Models}
\label{sec:background:zsrl}

\newcommand{\smath}[1]{\ensuremath{\small #1}}

BFMs are unsupervised RL agents based on successor features \cite{agarwal2025a, di2025unified}. The idea is to learn a latent-conditioned policy $\pi(a \mid s, z)$ from a reward-free dataset and infer the $z$ that maximizes a particular task of interest from a reward-labeled dataset, without further training \cite{jeen2025zero}. In general, we let $\phi:\mathcal S\to\mathbb R^d$ define latent rewards $\smath{r_z(s)=\phi(s)^\top z}$ and learn successor features $\smath{\psi(s,a,z)\approx \mathbb E_{\pi_z}[\sum_{t\ge 0}\gamma^t\phi(s_{t+1}) \mid s_0=s,a_0=a]}$, so that $\smath{Q_{\mathrm{SF}}(s,a,z)=\psi(s,a,z)^\top z}$ can serve as the critic in a standard off-policy actor-critic update \cite{lillicrap2015continuous} over some distribution of $z$, e.g.,
$
\small
\mathcal L_{\mathrm{actor}}
=
-
\mathbb E_{s\sim\mathcal D, z\sim p(z), a\sim\pi(\cdot\mid s,z)}
\left[
Q_{\mathrm{SF}}(s,a,z)
\right]
$.
In this work, the details of training $\psi$ and $\phi$ will be interchangeable and are not necessary to follow the main text, so we defer a more complete introduction to Appendix \ref{app:background:zsrl}. Most recent work is based on the FB algorithm \cite{touati2021learning, touati2022does}, but there are many alternatives and extensions \cite{chua2024learning, bagatella2026soft, zheng2026can, jeen2024zerolow, agarwal2025proto, park2024foundation}. Our experiments primarily use TD-JEPA \cite{bagatella2025td} because it shares similarities with off-policy supervised RL and lets us transfer implementation details from an optimized meta-RL baseline when appropriate.

BFMs benefit from diverse training datasets with broad state-action coverage, and are most commonly applied to offline datasets originally collected by a policy that optimized for exploration \cite{burda2018exploration, laskin2021urlb}. They adapt to a test-time task by fitting $\smath{z^*=\arg\min_z \sum_{\mathcal{D}_r}(r_i-\phi(s_i)^\top z)^2}$ on a reward-labeled dataset to deploy $\pi(\cdot\mid s,z^*)$, and we will refer to this inference process as zero-shot RL (ZSRL). Meta-RL's inference of $c_r$ plays a similar role as ZSRL's inference of $z^*$: both parameters estimate the active reward function and therefore enable generalization to novel objectives. ZSRL relies on an existing \textit{offline} dataset $\mathcal{D}_r$ to fit $z^*$, after which this parameter is fixed. In contrast, meta-RL typically infers $c_r$ from $\tau_{:t}$ (an \textit{online} $\mathcal{D}_r$ it has collected itself) by updating a belief between timesteps or episodes \cite{rakelly2019efficient}. This discrepancy makes ZSRL and meta-RL difficult to compare directly. In fact, ``zero-shot RL'' is, by design, unsuited to zero-shot generalization as typically defined (Section \ref{sec:background:metarl}), where we expect a policy to perform well in a novel task on the first attempt with no prior data. OptiBFM \cite{rupf2025optimistic} helps to close this gap by demonstrating that, as in meta-RL, we can maintain a belief over $z^*$ that is updated between timesteps or episodes. 

\begin{figure}[h!]
    \centering
    \vspace{-4mm}
    \includegraphics[width=0.80\linewidth]{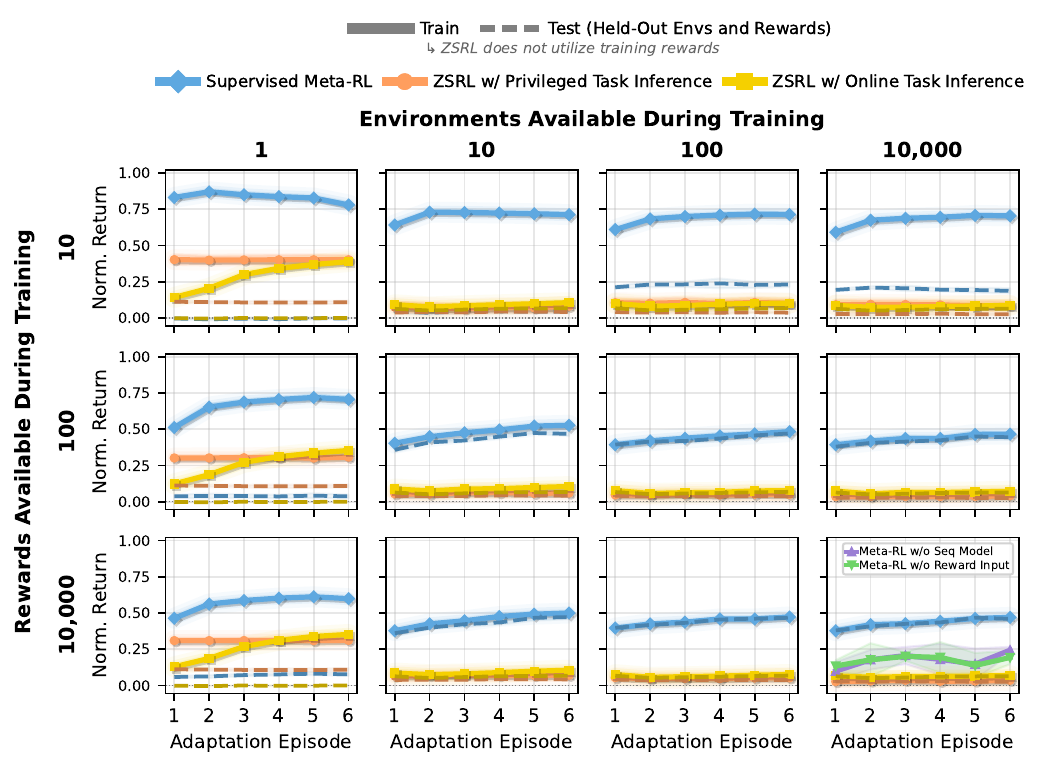}
    \vspace{-2mm}
\caption{\textbf{BFM and Meta-RL Generalization.} In offline ColorGrid ($n=9$), we vary training environments (columns) and rewards (rows), then evaluate $k=6$ episode adaptation on train and held-out tasks. Meta-RL improves with training set diversity but needs many reward functions to generalize; BFMs adapt in the fixed-environment setting (left column), but fail across environments.}
    \label{fig:colored_grid_9x9}
\end{figure}

\subsection{Generalization of BFMs and Supervised Meta-RL}

With OptiBFM's \textit{online task inference} enabled, BFMs and meta-RL can be directly compared. We will use a didactic ColorGrid problem as a running example through Section \ref{sec:method}. ColorGrid creates an $n \times n$ continuous 2D gridworld where the agent receives a reward based on the unit square cell of its current location, and the objective is to navigate to the cell with the highest reward and remain there. The underlying cell layout can be randomly permuted such that any two cells may be adjacent. In short, we can generate $(n^2)!$ distinct environments where nearly every transition is unpredictable without a reasonable estimate of the active grid permutation, $c_d$. Randomly sampling a reward for each cell generates infinitely many $c_r$ values. Additional environment details are in Appendix \ref{app:additional_envs}.

As a first experiment, we split $c_d$ and $c_r$ into train/test sets and generate tasks from their cross product, measuring environment and reward generalization separately. To replicate BFMs' high-coverage offline setting, we collect large random-policy datasets. Our BFM baseline is the better of FB and TD-JEPA, while our meta-RL baseline is an off-policy variant of RL$^2$ \cite{duan2016rl} that makes offline training reasonable, if still slightly unfair---we return to this later. Figure \ref{fig:colored_grid_9x9} reports results for $n=9$. Meta-RL follows a predictable trend: adding environments or reward functions improves generalization, but closing the train/test gap requires a diverse mixture of both. In this sense, supervised meta-RL is designed for the bottom-right corner of Fig. \ref{fig:colored_grid_9x9}. OptiBFM allows BFMs to match the standard $z^*$ inference procedure, though adaptation barely saturates within the $k=6$ test episodes. In contrast, meta-RL has trained on reward functions and is incentivized to learn a narrow prior over $c_r$. For now, we are not concerned with the performance gap between BFMs and meta-RL where BFMs show meaningful adaptation. Instead, we focus on BFMs' relative performance: the train curves in the left column of Fig. \ref{fig:colored_grid_9x9} represent the established BFM setting, and performance holds as evaluated reward functions grow in a fixed environment, even though the reward labels go unused and therefore act as a test set themselves. However, BFMs immediately collapse with multiple training environments, a problem also identified by \citet{bobrin2025zero}.

\section{BFMs as Meta-RL Agents}
\label{sec:method}

BFMs can adapt to unseen reward functions over $k$ attempts without train-time rewards. However, to be useful in meta-RL, they must also generalize across environments. This is only one of several gaps: meta-RL benefits from online training, while BFMs default to offline datasets; meta-RL manages exploration and exploitation over $k$ test episodes, while BFMs do not; and complete reward-free learning is often unnecessary in simulated online RL domains. We can usually define some reward-labeled tasks, even if we cannot cover every behavior needed at deployment. We investigate these gaps before proposing a solution that enables a competitive hybrid between meta-RL and BFMs.

\vspace{-2mm}

\begin{figure}[h!]
    \centering
    \begin{minipage}[b]{0.45\linewidth}
        \centering
        \includegraphics[width=\linewidth]{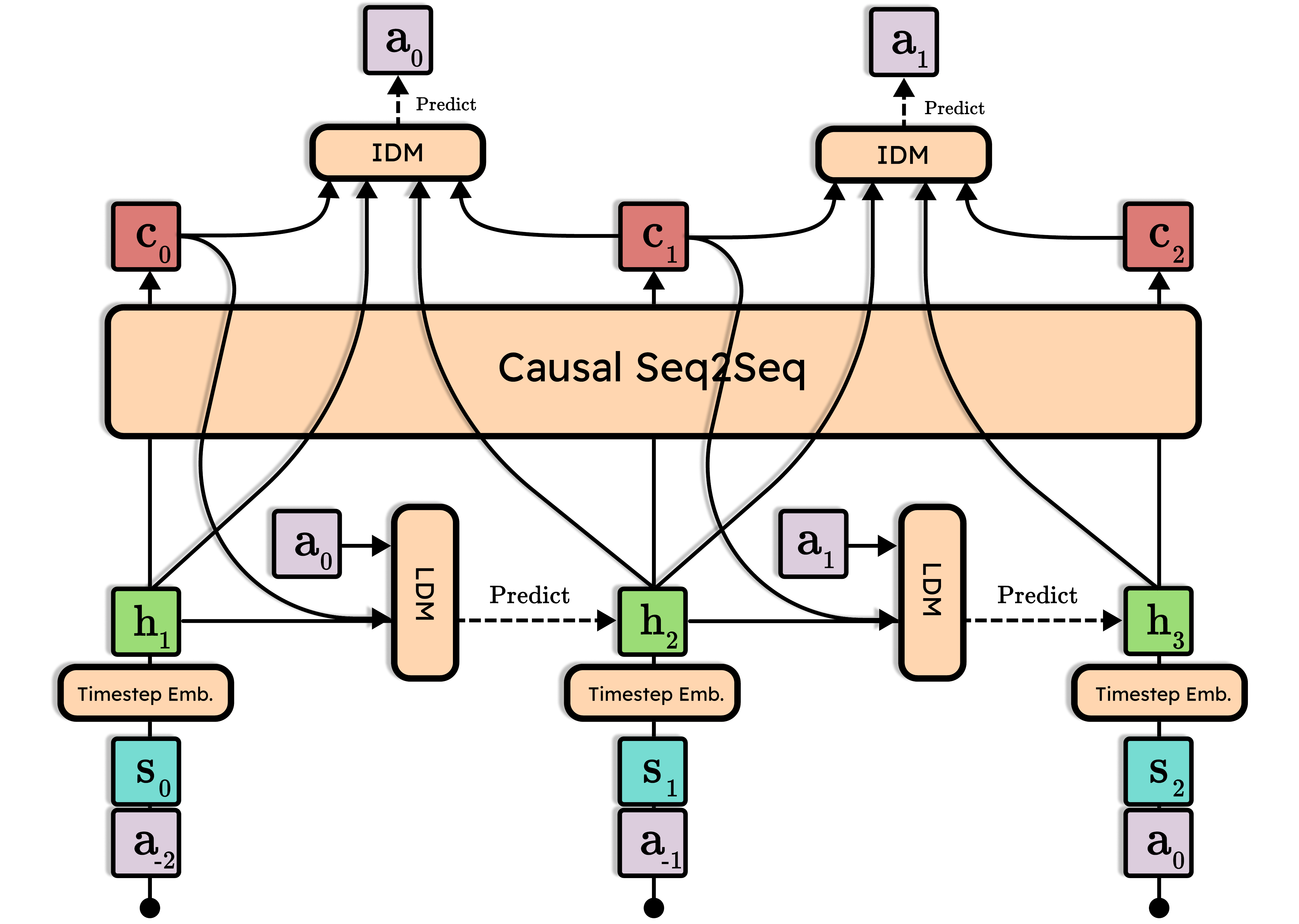}
    \end{minipage}
    \hfill
    \begin{minipage}[b]{0.40\linewidth}
        \centering
        \includegraphics[width=\linewidth]{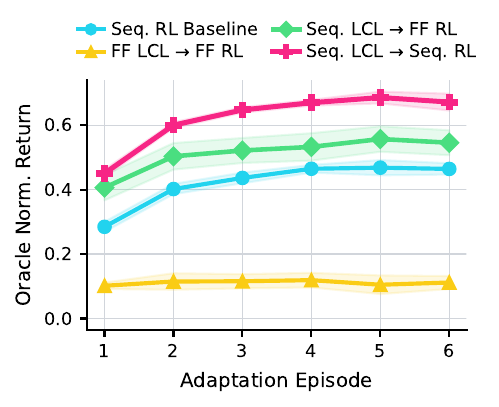}
    \end{minipage}
    \vspace{-2mm}
    \caption{\textbf{Latent Context Learning.} \textbf{(Left)} We learn a timestep embedding \hfig{h_t} and sequence model \sfig{S_\theta(\tau_{:t})} that generates an online estimate of the environment parameters, \cfig{\tilde{c}_t}. The representation is trained with self-supervised inverse and latent dynamics objectives; Algorithm \ref{alg:lcl} gives the full procedure. Action inputs \afig{a_t} are necessary context to infer dynamics, but are time-shifted to avoid leaking the IDM label. \textbf{(Right)} Context features learned by a feedforward (FF) or Transformer (Seq.) $S_{\theta}$ are presented to either a feedforward or sequence-based supervised RL policy $\pi$ in ColorGrid.}
    \label{fig:ldl_and_metarl_dynamics_adaptation}
\end{figure}

\subsection{Adapting to Novel Dynamics}

\label{sec:method:adapating}

An RL policy that cannot condition $\pi$ on estimates of the current dynamics, $c_d$, and reward function, $c_r$, learns an average policy under $p(c_d,c_r)$ that is generally not optimal for any specific task \cite{ghosh2019learning}. BFMs share this problem, but $z^*$ already plays the role of $c_r$, leaving us to infer $c_d$. Following context-based meta-RL \cite{beck2023survey}, we augment each trajectory $\tau$ with a learned estimate $\tilde{c}_d$ and train $\pi(a \mid s,\tilde{c}_d)$. Because context labels are assumed to be unavailable, we use a self-supervised dynamics objective. Unlike BeliefFB \cite{bobrin2025zero}, we target timestep-level online task inference with latent dynamics representations that avoid method-specific changes to the BFM update.

\renewcommand{\hfig}[1]{\textcolor{figgreen}{\ensuremath{#1}}}

\begin{figure}[h!]
    \centering
    \includegraphics[width=\linewidth]{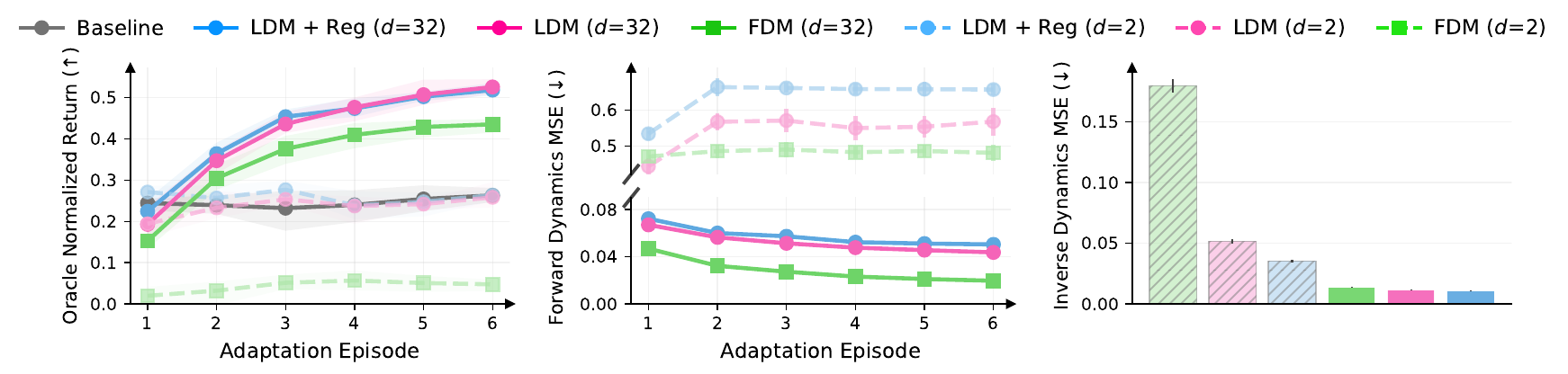}
    \vspace{-5mm}
    \caption{\textbf{LCL and Generalization.} TD-JEPA in ColorGrid $n=5$ with various context embedding sizes and training objectives. Performance correlates with lower FDM and IDM error. LDM representations recover similar dynamics prediction as FDM without training on this objective.}
    \label{fig:lcl_metrics}
\end{figure}

Figure \ref{fig:ldl_and_metarl_dynamics_adaptation} shows our Latent Context Learning (LCL) architecture, which produces $\tilde{c}_d$ at every timestep during rollouts. We embed each timestep of the trajectory $\tau_{:t}$ to create the input sequence $\{h_0,\dots,h_t\}$, then pass it through a causal sequence model. Let $S_\theta(\tau_{:t}) \in \mathbb{R}^d$ be the output representation at timestep $t$. This output is trained to identify the active dynamics through self-supervised inverse dynamics modeling (IDM) and latent dynamics modeling (LDM). Because accurate estimates of $c_d$ improve these objectives, and because $S_{\theta}$ is the only representation with access to the necessary history, we treat $\tilde{c}_t=S_{\theta}(\tau_{:t})$ and can train a standard BFM on augmented states $\bar{s}_t=(s_t,\tilde{c}_t)$. $S_{\theta}$ observes action choices to infer dynamics, but we avoid leaking the inverse dynamics model label by staggering the action input by two timesteps. We prefer latent dynamics modeling because forward dynamics modeling (FDM) scales with the size of the state space. Following latent world modeling and self-supervised sequence models \cite{oord2018representation, schwarzer2020data, schwarzer2023bigger}, we use an EMA target network for $h_{t+1}$ labels \cite{grill2020bootstrap}, and can optionally encourage $\tilde{c}_t$ to be dissimilar across sequences in each training batch \cite{jajoo2026regularized}. Appendix \ref{app:implementation:lcl} and Algorithm \ref{alg:lcl} provide additional details. By default, we train IDM and LDM jointly; detached FDM prediction error is used only as a diagnostic metric.

\vspace{-1mm}

\begin{figure}[h!]
    \centering
    \includegraphics[width=.97\linewidth]{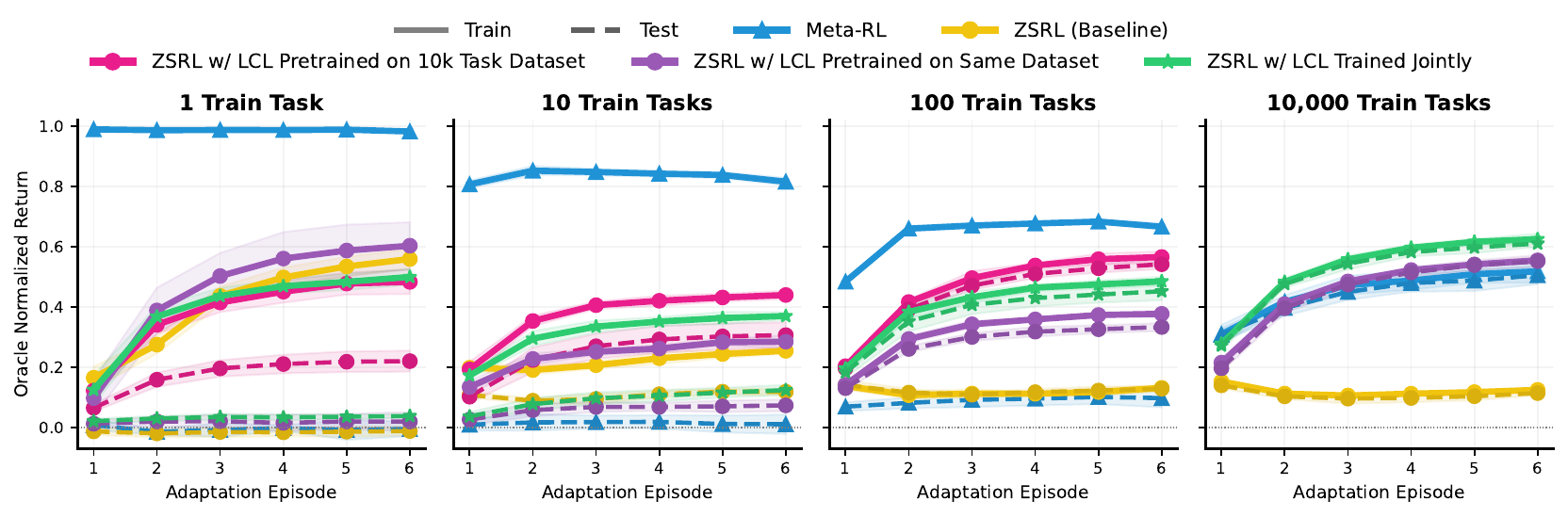}
    \vspace{-3mm}
    \caption{\textbf{Latent context learning improves BFM generalization.} We increase the number of training $(c_d,c_r)$ task pairs with separate offline datasets. Standard TD-JEPA collapses across environments, while LCL-augmented TD-JEPA recovers adaptation and improves test performance as the training set grows. Pretraining LCL is most helpful at small training set sizes, but joint training approaches the same performance at scale.}
    \label{fig:lcl_offline}
\end{figure}

Context estimates should improve supervised RL by reducing or eliminating the need for long-term memory policies, and Figure \ref{fig:ldl_and_metarl_dynamics_adaptation} (Right) verifies this effect in ColorGrid. By augmenting states with context estimates, we can also make BFM networks context-aware without the instability of sequence models. Figure \ref{fig:lcl_metrics} shows that, when the context representation $S_{\theta}(\tau) \in \mathbb{R}^{T\times d}$ has enough capacity to represent $c_d$, joint IDM and LDM learns accurate dynamics estimates that improve TD-JEPA performance. In Figure \ref{fig:lcl_offline}, we define train/test splits of $(c_d,c_r)$ task pairs with separate offline datasets, and evaluate generalization as the number of training tasks grows. Since LCL and the downstream BFM agent are independent, they can be trained separately. Pretraining and freezing LCL on the largest task set improves BFM generalization at small training set sizes, but gives no clear advantage over training both systems from scratch on the same offline dataset. This is encouraging because joint training is the only realistic option when the dataset is collected online.

To further illustrate BFMs' failure across environments and test whether context learning addresses it, we turn to motion tracking for the G1 humanoid, building on BFM-Zero \cite{li2025bfm}. We randomize the dynamics through wind and gravity perturbations of varying strength, and evaluate tracking performance on the LAFAN dataset \cite{harvey2020robust}, which contains $4.6$ hours of motion sequences. We compare a BFM trained without these perturbations (BFM-Single) to one trained across the full range of dynamics (BFM-Multi), with and without the FDM and LDM context-learning objectives. Consistent with our analysis above, a BFM trained across multiple dynamics averages its actions over them, which can degrade performance: in the in-distribution setting, BFM-Multi performs even worse than BFM-Single, suggesting that BFMs suffer substantially when they cannot infer the environment's dynamics. Context learning recovers much of this gap (Table \ref{tab:humanoid}), suggesting its promise for training across domains. Surprisingly, it also improves performance modestly in the out-of-distribution setting. Appendix \ref{app:humanoid} illustrates these BFM failures and provides additional details on the task and environment.

\begin{table*}[t]
\centering
\small
\caption{\textbf{BFMs with latent context learning improve humanoid motion tracking.} We compare motion-tracking metrics for humanoid locomotion under in-distribution and out-of-distribution dynamics. BFM-Single trains without wind/gravity randomization and can be viewed as an in-distribution oracle for BFM-Multi, or as a worst-case baseline under OOD dynamics.}
\label{tab:humanoid}
\resizebox{\textwidth}{!}{
\begin{tabular}{lcccccccc}
\toprule
\multirow{2}{*}{Method} 
& \multicolumn{4}{c}{In-Distribution} 
& \multicolumn{4}{c}{Out-of-Distribution} \\
\cmidrule(lr){2-5} \cmidrule(lr){6-9}
& mpjpe-l($\downarrow$) & vel. dist.($\downarrow$) & accel. dist.($\downarrow$) & emd($\downarrow$)
& mpjpe-l($\downarrow$) & vel. dist.($\downarrow$) & accel. dist.($\downarrow$) & emd($\downarrow$) \\
\midrule
BFM-Single 
& 117.2 & 1.64 & 2.75 & 1.01
& 194.7 & 2.81 & 4.83 & 1.80 \\
\cmidrule(lr){1-9}
BFM-Multi
& 132.7 & 1.77 & 2.96 & 1.13
& 151.6 & 2.13 & 3.63 & 1.39 \\

FDM+BFM-Multi 
& 125.3 & 1.73 & 2.90 & 1.09
& 149.3 & 2.09 & 3.54 & 1.37 \\

LDM+BFM-Multi 
& \bf 121.5 & \bf 1.68 & \bf 2.85 & \bf 1.05
& \bf 145.9 & \bf 2.06 & \bf 3.52 & \bf 1.36 \\
\bottomrule
\end{tabular}
}
\end{table*}

\subsection{Exploring, Exploiting, and Learning Online}
\label{sec:method:online}

BFM algorithms are primarily developed on offline RL benchmarks \cite{laskin2021urlb} with behavior datasets generated by an online policy optimizing an intrinsic reward \cite{burda2018exploration,  seo2021state, Raileanu2020RIDE:}. BFMs rely on high state-action coverage and struggle to learn online because the actor objective $\mathcal{L}_{\mathrm{actor}}$ collapses onto narrow behavior. Appendix \ref{app:additional_results:online} explores the difficulty of online learning and task inference in a standard locomotion benchmark. DVFB \cite{sun2025unsupervised} addresses dataset diversity by maximizing a combination of $Q_{\mathrm{SF}}$ and a supervised critic trained on intrinsic curiosity rewards, which we will call $Q_R$ (Equation \ref{eq:dvfb}). Recent applications of BFMs to humanoid robotics \cite{tirinzonizero, li2025bfm} use a similar idea, but derive $Q_R$ rewards from behavior in an expert motion-capture dataset.

\vspace{-4mm}

\begin{align}
\mathcal{L}_{\mathrm{DVFB}}
&= -\,\mathbb{E}_{s\sim\mathcal{D},\, z\sim p(z),\, a\sim\pi(\cdot\mid s,z)}
\!\bigl[\,Q_{\mathrm{SF}}(s,a,z) \;+\; \alpha\, Q_{R}(s,a,z)\,\bigr].
\label{eq:dvfb}
\end{align}

In meta-RL problems, we can take this idea further. First, we replace the \textit{global} intrinsic reward---one that decreases as a function of total training steps, such as RND \cite{burda2018exploration}---with an \textit{episodic} reward that decays within a $k$-episode meta-trial \cite{henaff2023study}. An agent maximizing this reward is incentivized to explore new states across each episode at test time. We use an intrinsic reward that combines an off-the-shelf episodic bonus (NGU \cite{badia2020never} or E3B \cite{henaff2022exploration}) on LCL timestep features $h_t$ (Fig. \ref{fig:ldl_and_metarl_dynamics_adaptation}) with the LCL prediction error from the previous timestep. The latter is a core meta-RL idea \cite{zintgraf2021exploration, zhang2021metacure}: it incentivizes exploration that improves the $\tilde{c}_d$ estimate, which can also improve in-context. Second, because episodic rewards are non-Markov, $Q_R$ and $\pi$ should have memory and take LCL-augmented trajectories $\bar{\tau}$ as input. Finally, we sample $\alpha$ during training and condition $Q_R$ on the active value:

\vspace{-4mm}

\begin{align}
\mathcal{L}_{\alpha\text{-}\mathrm{BFM}}
&= -\,\mathbb{E}_{\substack{\textcolor{hltau}{\tau}\sim\mathcal{D},\; z\sim p(z),\\ \textcolor{hlalpha}{\alpha}\sim p(\textcolor{hlalpha}{\alpha}),\; a\sim\pi(\cdot\mid \textcolor{hltau}{\bar{\tau}},\,z,\,\textcolor{hlalpha}{\alpha})}}
\!\Bigl[\,(1-\textcolor{hlalpha}{\alpha})\, Q_{\mathrm{SF}}(\bar{s},a,z) \;+\; \textcolor{hlalpha}{\alpha}\, Q_{R}(\textcolor{hltau}{\bar{\tau}},a,z,\textcolor{hlalpha}{\alpha})\,\Bigr].
\label{eq:alphabfm}
\end{align}

\begin{figure}[h!]
    \centering
    \includegraphics[width=\linewidth]{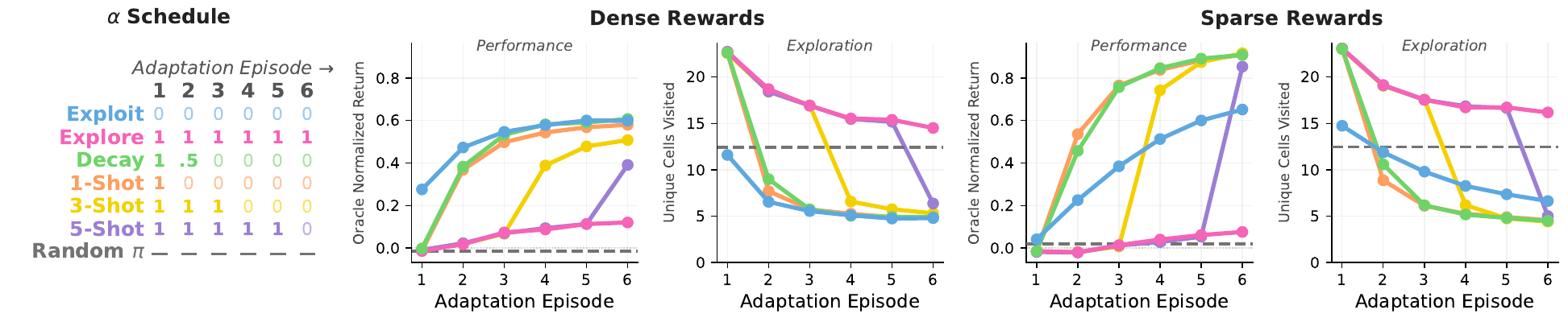}
    \vspace{-4mm}
    \caption{\textbf{Exploring at Test Time.} ColorGrid ($n=5$) policies trained by Eq. \ref{eq:alphabfm} can vary $\alpha$ over $k=6$ test episodes. High $\alpha$ explores more of the state space, improving online task estimates, which can then be exploited to outperform the standard BFM policy ($\alpha=0$, blue) when rewards are sparse.}
    \label{fig:alpha_schedule_demo}
\end{figure}

The result is a policy that generates more diverse data during training while allowing $\alpha$ to vary at test time. Due to our choice of intrinsic reward, $\alpha=1$ drives the policy to explore new behavior each episode in hopes of (1) finding rewards that improve OptiBFM's online task inference of $z^*$ and (2) improving the context estimate $\tilde{c}_d$. In contrast, $\alpha=0$ exploits that knowledge to maximize returns under $z^*$. We can schedule $\alpha$ to recover meta-RL's $k$-shot exploration-exploitation tradeoff at test time, as demonstrated in Figure \ref{fig:alpha_schedule_demo}.

\subsection{Learning from Available Rewards}
\label{sec:rews}

BFMs assume access to zero reward labels during training. While this may be necessary when learning offline, such as from a web dataset that was not intended for RL, the constraint is unnecessary when learning online in simulated environments. When we create meta-RL or multi-task RL training domains, we can usually create some reward-labeled tasks; the challenge is creating enough to cover all behaviors we may want during deployment. Adding training tasks should improve generalization with diminishing returns, creating a clear path to improving the agent by improving its dataset. Ideally, we would bring this relationship to BFMs. Each new train-time objective can help span the space of possible behaviors and supervise learning progress, while unsupervised learning fills in gaps and finds solutions to objectives we had not considered.

We can approach this problem from two directions. The first is to let reward labels improve the BFM update itself. For example, BFMs learn features $\phi(s)$ and assume $\phi(s)^\top z = r$; when reward labels are available during training, we can enforce this relationship directly with an auxiliary loss. It has also become standard to sample training $z \sim p(z)$ from recent values of $\phi(s)$, implicitly assuming evaluation on sparse state-based goal-reaching \cite{touati2021learning, touati2022does}. With reward labels, we can extend this idea to arbitrary reward functions by using online task inference along the labeled trajectory to estimate $z^*$, as if the same states and rewards appeared at test time. Appendix \ref{app:additional_results:rewards} evaluates both ideas in ColorGrid and DMC ExoRL, finding that they make $\phi$ more predictive of rewards but have mixed effects on policy performance. A second approach is to restore supervised meta-RL where rewards are available and create a true hybrid method. We modify the ``intrinsic'' critic $Q_R$ from Section \ref{sec:method:online} to learn from a weighted combination of intrinsic rewards, $r_i$, and extrinsic (supervised) rewards $r_e$:

\vspace{-4mm}

\begin{align}
Q_{R}(\textcolor{hltau}{\bar{\tau}},a,z,\textcolor{hlalpha}{\alpha},\textcolor{hlbeta}{\beta})
&\approx
\underbrace{%
  (1-\textcolor{hlbeta}{\beta})\,r_{i}
  + \textcolor{hlbeta}{\beta}\,r_{e}
}_{\mathclap{\text{Intrinsic and Extrinsic Reward}}}
+ \gamma\,
Q_{R}\!\bigl(\textcolor{hltau}{\bar{\tau}'},a',z,\textcolor{hlalpha}{\alpha},\textcolor{hlbeta}{\beta}\bigr)
\label{eq:QI-bellman}
\end{align}

where $\bar{\tau}'$ and $a'$ are the next timestep in the standard TD backup. We are then free to repeat the idea from Equation \ref{eq:alphabfm} of sampling the weight $\beta$ during training and conditioning the actor and critic on the current value so that it may also be adjusted at deployment:

\vspace{-4mm}

\begin{align}
\mathcal{L}_{\alpha\beta\text{-}\mathrm{BFM}}
&= -\,\mathbb{E}_{\substack{\textcolor{hltau}{\tau}\sim\mathcal{D},\; z\sim p(z),\\ \textcolor{hlalpha}{\alpha}\sim p(\textcolor{hlalpha}{\alpha}),\; \textcolor{hlbeta}{\beta}\sim p(\textcolor{hlbeta}{\beta}),\\ a\sim\pi(\cdot\mid \textcolor{hltau}{\bar{\tau}},\,z,\,\textcolor{hlalpha}{\alpha},\,\textcolor{hlbeta}{\beta})}}
\!\Bigl[\,(1-\textcolor{hlalpha}{\alpha})\, Q_{\mathrm{SF}}(\bar{s},a,z)
  \;+\; \textcolor{hlalpha}{\alpha}\, Q_{R}(\textcolor{hltau}{\bar{\tau}},a,z,\textcolor{hlalpha}{\alpha},\textcolor{hlbeta}{\beta})\,\Bigr]
\label{eq:alphabetabfm}
\end{align}

\begin{wrapfigure}{r}{0.42\linewidth}
    \centering
    \vspace{-10mm}
    \includegraphics[width=\linewidth]{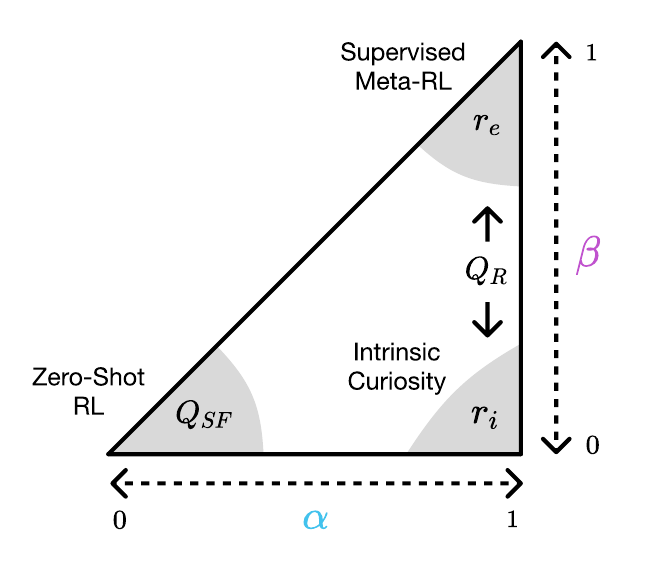}
    \vspace{-9mm}
    \caption{\textbf{A Unified Objective.} The $\mathcal{L}_{\alpha\beta\text{-}\mathrm{BFM}}$ loss (Eq. \ref{eq:alphabetabfm}) interpolates between zero-shot RL, intrinsic curiosity, and supervised meta-RL. $\alpha$ and $\beta$ can be configured to adjust generalization behavior.}
    \label{fig:alpha_beta_compass}
    \vspace{-2mm}
\end{wrapfigure}

When $\beta=0$, we recover the exploratory ZSRL objective of Eq. \ref{eq:alphabfm}, and when $\alpha=\beta=0$, we recover standard ZSRL. Most importantly, when $\alpha=\beta=1$, we recover supervised meta-RL. In practice, this makes it possible to compete with a standard supervised meta-RL baseline on in-distribution tasks, while opening the possibility of generalizing via $z$ to novel tasks when meta-RL fails. Figure \ref{fig:alpha_beta_compass} visualizes this space of possible test-time settings.

\section{MetaBFM}
\label{sec:metabfm}

We can now combine the pieces from Section \ref{sec:method} into a unified update, \texttt{MetaBFM}, that captures the in-distribution performance of supervised meta-RL and the out-of-distribution generalization of unsupervised BFMs. At each training step, we first update the Latent Context Learning module with its self-contained objective (Algorithm \ref{alg:lcl}). LCL's primary role is to augment each state with a context estimate $\tilde{c}_d$, allowing the BFM networks to learn from data generated by multiple environment dynamics (Section \ref{sec:method:adapating}). LCL also produces per-timestep representations that can be passed to off-the-shelf episodic curiosity methods to compute intrinsic rewards $r_i$ for the trajectory. We then update the BFM networks, excluding the actor $\pi$, over a distribution $p(z)$. Any actions from $\pi$ required for the BFM update are sampled from $\pi(\cdot \mid \bar{\tau}, z, \alpha=0, \beta=0)$, which recovers the original BFM policy; the remaining details are unchanged, and many ZSRL algorithms are interchangeable here, though we use TD-JEPA. Next, we update the critic head $Q_R$ with a standard off-policy backup on weighted intrinsic and extrinsic rewards (Equation \ref{eq:QI-bellman}) and compute the actor loss $\mathcal{L}_{\alpha\beta\text{-}\mathrm{BFM}}$ (Equation \ref{eq:alphabetabfm}), both over samples from $p(\alpha,\beta)$. We collect trajectories from the online environment by sampling from the same distribution of $z$, $\alpha$, and $\beta$.

While this update is more expensive than training the BFM component in isolation, in practice we share network backbones for efficiency. For example, $\pi$ and $Q_R$ share a Transformer backbone to enable long-term memory. Many such training details are copied directly from a strong off-policy meta-RL implementation \cite{grigsby2024amago}, and Appendix \ref{app:implementation} provides additional discussion. To make extrinsic rewards optional, we force $\beta=0$ when collecting or training on reward-free trajectories. Early layers that take extrinsic reward inputs to $Q_R$ and $\pi$ have their outputs multiplied by $\beta$, allowing us to insert and automatically ignore an arbitrary placeholder reward.

\begin{figure}[h!]
    \centering
    \includegraphics[width=0.49\linewidth]{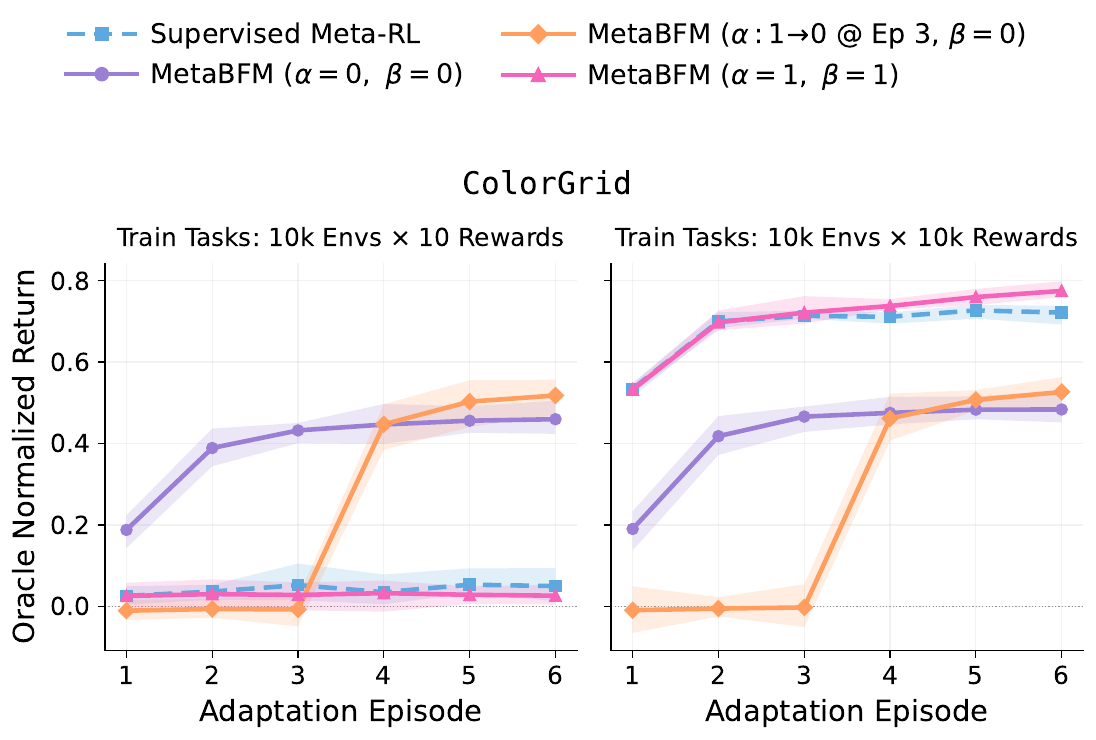}
    \hfill
    \includegraphics[width=0.49\linewidth]{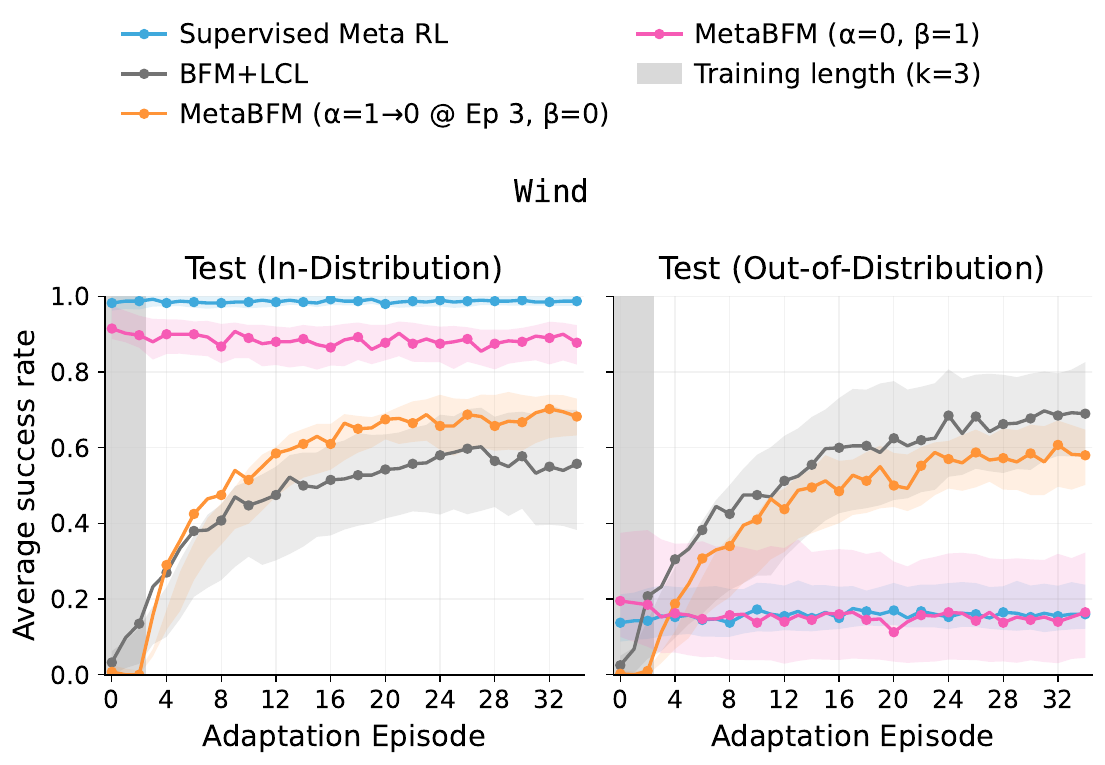}
    \vspace{-1mm}
    \caption{\textbf{Configurable MetaBFM Inference.} (\textbf{Left}) When learning from limited training objectives, \texttt{MetaBFM} in ZSRL mode ($\beta=0$) avoids supervised meta-RL's overfitting. (\textbf{Center Left}) With more objectives, supervised mode recovers the meta-RL baseline. (\textbf{Center Right}) On held-out in-distribution Wind tasks, \texttt{MetaBFM} remains competitive with supervised meta-RL. (\textbf{Right}) On out-of-distribution Wind tasks, ZSRL-style inference adapts over long adaptation horizons.}
    \label{fig:configurable_metabfm}
\end{figure}

Figure \ref{fig:colored_grid_9x9} showed that BFMs can learn adaptive behavior from limited training domains where supervised meta-RL overfits, but also that meta-RL improves with diverse datasets while BFMs do not. \texttt{MetaBFM}'s main goal is to allow for test-time configuration to take advantage of both of these methods' strengths while being able to avoid their weaknesses. Figure \ref{fig:configurable_metabfm} Left returns to this experiment and shows that \texttt{MetaBFM} captures both modes: a single policy can be deployed in ``ZSRL mode'' ($\alpha=0,\beta=0$) when meta-RL fails to generalize, or in ``reward-following mode'' ($\alpha=1,\beta=1$) when the dataset is diverse enough to support it.

\vspace{-2mm}

An extension of learning more general behavior from fewer train-time objectives is generalizing to test-time objectives that are clearly out of distribution from the supervised signals seen during training. We study this with a variant of the Wind environment \cite{dorfman2020offline, ni2022recurrent}, a common benchmark for OOD generalization. Agents navigate from the origin toward a randomly placed goal, $c_r$, while adjusting for wind in a random direction, $c_d$. The training set contains all wind directions but excludes goals in the upper-left quadrant. At test time, we therefore evaluate both held-out in-distribution tasks (any wind direction, goals in the three training quadrants) and out-of-distribution tasks (any wind direction, goals in the excluded quadrant). Figure \ref{fig:configurable_metabfm} Right shows the adaptation curves for each set. \texttt{MetaBFM} significantly outperforms the supervised baseline on the OOD test set when configured to maximize successor-feature values, while nearly matching its performance on in-distribution tasks where train-time rewards are useful. Perhaps the most interesting OOD result is the gap between the train-time trajectory length of $k=3$ episodes and the length at which online task inference saturates ($k>20$). This kind of length extrapolation is rare in black-box meta-RL and may be enabled by offloading the usual context estimation of $c_r$ outside of the policy weights and into ZSRL's $z^*$.

\begin{figure}[h!]
    \vspace{-2mm}
    \centering
    \includegraphics[width=\linewidth]{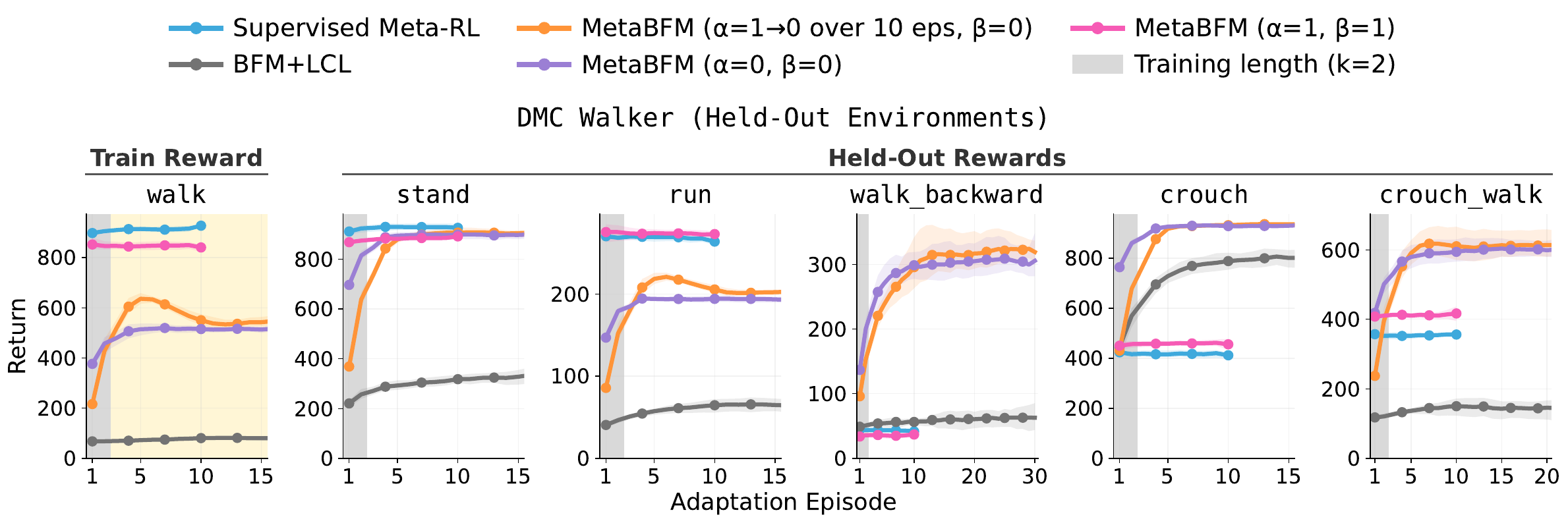}
    \vspace{-6mm}
    \caption{\textbf{DMC Walker.} Adaptation in held-out environments after training on $10$k environments with only the \emph{walk} reward (left). \texttt{MetaBFM} recovers supervised meta-RL on the train reward and its close variants, and generalizes to novel rewards in ZSRL mode. ZSRL evaluations run for $30$ episodes; x-axes are truncated after performance plateaus.}
    \label{fig:dmc_walker}
\end{figure}

We repeat this experiment on the ``walker'' locomotion task in the DeepMind Control Suite \cite{tassa2018deepmind}, using a variant of the meta-RL setup from \citet{rothfuss2018promp}. We generate $c_d$ by randomizing body geometry, joint ranges, motor damping, and ground friction. Agents train over $k=2$ episodes in $10$k random walker variations using only the standard \emph{walk} reward, and are evaluated on held-out dynamics (Appendix \ref{app:dmc_walker}). At test time, we evaluate on \emph{walk}, two closely related objectives (\emph{stand} and \emph{run}), and three objectives requiring behaviors not encouraged during supervised training (\emph{walk backward}, \emph{crouch}, and \emph{crouch walk}). Figure \ref{fig:dmc_walker} shows the same pattern as Wind: on \emph{walk}, \texttt{MetaBFM} in reward-following mode performs similarly to supervised meta-RL, while ZSRL mode is noticeably weaker. On the three novel objectives, supervised meta-RL fails to follow the test-time reward signal, while performance with ZSRL-style inference continues to improve well past the training horizon.

\begin{figure}[h!]
    \vspace{-1mm}
    \centering
    \includegraphics[width=\linewidth]{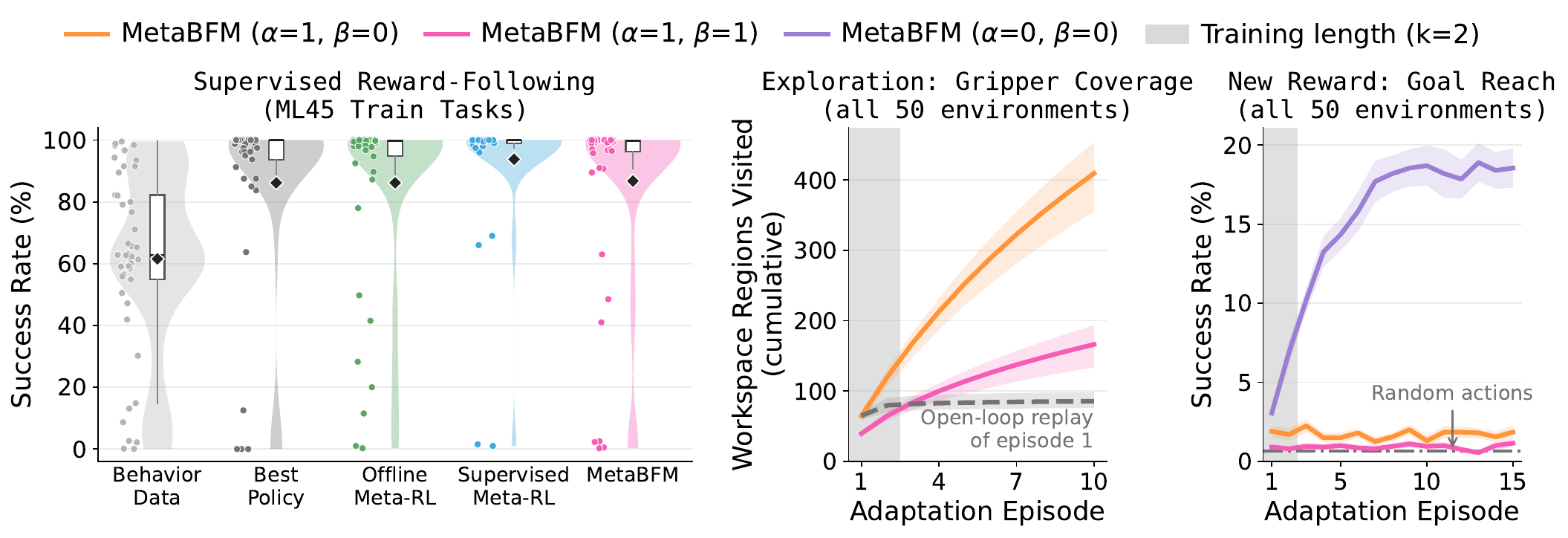}
    \vspace{-5mm}
  \caption{\textbf{MetaWorld ML45.} (Left) In reward-following mode, \texttt{MetaBFM} recovers supervised meta-RL performance on the $45$ training tasks. Boxes show interquartile ranges; diamonds show mean success rates. Behavior Data and Best Policy report success rates for the seed dataset and its best contributing policy, respectively. (Center) In exploration mode, \texttt{MetaBFM} continues to expand the gripper's coverage of the tabletop workspace over a trial. (Right) In ZSRL mode, \texttt{MetaBFM} adapts to a simple unseen goal-reaching reward across all $50$ environments.}
    \label{fig:ml45}
\end{figure}

Finally, we evaluate \texttt{MetaBFM} on the more complex MetaWorld continuous-control benchmark \cite{yu2020meta}, using the hardware-accelerated variant from MTBench \cite{joshi2025benchmarking}. Despite learning to maximize a range of actor objectives at both training and test time, \texttt{MetaBFM} converges to near-perfect performance on six ML1 tasks, each testing generalization across hidden-parameter variations of a single tabletop manipulation skill (Appendix Figure~\ref{fig:ml1}). Figure~\ref{fig:ml45} reports results for the full multi-task ML45 setting. To reduce computational cost, we initialize off-policy learning with a replay buffer of demonstrations collected from checkpoints of a multi-day meta-RL training run, with exploration noise enabled. We compare baselines trained offline on this dataset with meta-RL and \texttt{MetaBFM} agents that continue collecting online experience. On the training tasks, \texttt{MetaBFM} again performs comparably to supervised meta-RL (Figure~\ref{fig:ml45}, left). In exploration mode, \texttt{MetaBFM} continues to visit new regions of the workspace in each episode, while reward-following mode largely repeats its first attempt (Figure~\ref{fig:ml45}, center). However, \texttt{MetaBFM} fails to generalize to MetaWorld's five held-out manipulation tasks in either reward-following or ZSRL inference mode (Appendix Figure~\ref{fig:ml45_sigma}). To test adaptation to novel objectives that require in-distribution behaviors, we add a goal-reaching reward to all $50$ training and held-out environments, with hidden targets drawn from states visited during exploration rollouts (Appendix~\ref{app:metaworld}). Performance with ZSRL-style inference continues to improve well beyond the $k=2$ training horizon, while success rates in reward-following and exploration modes remain near chance (Figure~\ref{fig:ml45}, right).

\section{Conclusion}

Meta-RL requires diverse training reward functions, but anticipating every test-time objective remains difficult even as environments become cheaper to generate. Building on OptiBFM's online task inference, we extend BFMs to match a broader range of meta-RL's capabilities, including adaptation to novel environments. \texttt{MetaBFM} combines a self-predictive latent context for environment identification, intrinsic curiosity for online data collection and test-time exploration, and supervised value learning to take advantage of mixed reward supervision. The resulting policy recovers the in-distribution performance of supervised meta-RL while generalizing to reward functions outside the training set. Continued progress in meta-RL and BFMs could be mutually reinforcing, advancing generalist RL by shifting the emphasis from specifying additional training objectives to generating more diverse environments.

\begin{ack}
This work was supported by the National Science Foundation (CAREER Award FRR-2145283 and EFRI-2318065), the Office of Naval Research (N00014-24-1-2550), the Army Research Lab (W911NF-25-1-0065), the KIST-UT collaboration (UTAUS-FA00004578), the JP Morgan Chase Faculty Fellowship in AI, and the Sony Research Awards.
\end{ack}

\bibliography{citations}

@article{burda2018exploration,
  title={Exploration by random network distillation},
  author={Burda, Yuri and Edwards, Harrison and Storkey, Amos and Klimov, Oleg},
  journal={arXiv preprint arXiv:1810.12894},
  year={2018}
}

@article{hafner2023mastering,
  title={Mastering Diverse Domains through World Models},
  author={Hafner, Danijar and Pasukonis, Jurgis and Ba, Jimmy and Lillicrap, Timothy},
  journal={arXiv preprint arXiv:2301.04104},
  year={2023}
}

@article{team2023human,
  title={Human-Timescale Adaptation in an Open-Ended Task Space},
  author={Team, Adaptive Agent and Bauer, Jakob and Baumli, Kate and Baveja, Satinder and Behbahani, Feryal and Bhoopchand, Avishkar and Bradley-Schmieg, Nathalie and Chang, Michael and Clay, Natalie and Collister, Adrian and others},
  journal={arXiv preprint arXiv:2301.07608},
  year={2023}
}

@inproceedings{schwarzer2023bigger,
  title={Bigger, Better, Faster: Human-level Atari with human-level efficiency},
  author={Schwarzer, Max and Ceron, Johan Samir Obando and Courville, Aaron and Bellemare, Marc G and Agarwal, Rishabh and Castro, Pablo Samuel},
  booktitle={International Conference on Machine Learning},
  pages={30365--30380},
  year={2023},
  organization={PMLR}
}

@article{lu2023structured,
  title={Structured state space models for in-context reinforcement learning},
  author={Lu, Chris and Schroecker, Yannick and Gu, Albert and Parisotto, Emilio and Foerster, Jakob and Singh, Satinder and Behbahani, Feryal},
  journal={arXiv preprint arXiv:2303.03982},
  year={2023}
}

@misc{ni2022recurrent,
      title={Recurrent Model-Free RL can be a Strong Baseline for Many POMDPs}, 
      author={Tianwei Ni and Benjamin Eysenbach and Ruslan Salakhutdinov},
      year={2022},
      eprint={2110.05038},
      archivePrefix={arXiv},
      primaryClass={cs.LG}
}

@inproceedings{finn2017model,
  title={Model-agnostic meta-learning for fast adaptation of deep networks},
  author={Finn, Chelsea and Abbeel, Pieter and Levine, Sergey},
  booktitle={International conference on machine learning},
  pages={1126--1135},
  year={2017},
  organization={PMLR}
}

@inproceedings{yu2020meta,
  title={Meta-world: A benchmark and evaluation for multi-task and meta reinforcement learning},
  author={Yu, Tianhe and Quillen, Deirdre and He, Zhanpeng and Julian, Ryan and Hausman, Karol and Finn, Chelsea and Levine, Sergey},
  booktitle={Conference on robot learning},
  pages={1094--1100},
  year={2020},
  organization={PMLR}
}

@article{rothfuss2018promp,
  title={Promp: Proximal meta-policy search},
  author={Rothfuss, Jonas and Lee, Dennis and Clavera, Ignasi and Asfour, Tamim and Abbeel, Pieter},
  journal={arXiv preprint arXiv:1810.06784},
  year={2018}
}

@article{lillicrap2015continuous,
  title={Continuous control with deep reinforcement learning},
  author={Lillicrap, Timothy P and Hunt, Jonathan J and Pritzel, Alexander and Heess, Nicolas and Erez, Tom and Tassa, Yuval and Silver, David and Wierstra, Daan},
  journal={arXiv preprint arXiv:1509.02971},
  year={2015}
}

@article{tassa2018deepmind,
  title={Deepmind control suite},
  author={Tassa, Yuval and Doron, Yotam and Muldal, Alistair and Erez, Tom and Li, Yazhe and Casas, Diego de Las and Budden, David and Abdolmaleki, Abbas and Merel, Josh and Lefrancq, Andrew and others},
  journal={arXiv preprint arXiv:1801.00690},
  year={2018}
}

@article{ghosh2019learning,
  title={Learning to reach goals via iterated supervised learning},
  author={Ghosh, Dibya and Gupta, Abhishek and Reddy, Ashwin and Fu, Justin and Devin, Coline and Eysenbach, Benjamin and Levine, Sergey},
  journal={arXiv preprint arXiv:1912.06088},
  year={2019}
}

@inproceedings{rakelly2019efficient,
  title={Efficient off-policy meta-reinforcement learning via probabilistic context variables},
  author={Rakelly, Kate and Zhou, Aurick and Finn, Chelsea and Levine, Sergey and Quillen, Deirdre},
  booktitle={International conference on machine learning},
  pages={5331--5340},
  year={2019},
  organization={PMLR}
}

@article{zintgraf2021varibad,
  title={VariBAD: Variational Bayes-Adaptive Deep RL via Meta-Learning},
  author={Zintgraf, Luisa and Schulze, Sebastian and Lu, Cong and Feng, Leo and Igl, Maximilian and Shiarlis, Kyriacos and Gal, Yarin and Hofmann, Katja and Whiteson, Shimon},
  journal={Journal of Machine Learning Research},
  volume={22},
  number={289},
  pages={1--39},
  year={2021}
}

@article{dorfman2020offline,
  title={Offline meta learning of exploration},
  author={Dorfman, Ron and Shenfeld, Idan and Tamar, Aviv},
  journal={arXiv preprint arXiv:2008.02598},
  year={2020}
}

@inproceedings{liu2021decoupling,
  title={Decoupling exploration and exploitation for meta-reinforcement learning without sacrifices},
  author={Liu, Evan Z and Raghunathan, Aditi and Liang, Percy and Finn, Chelsea},
  booktitle={International conference on machine learning},
  pages={6925--6935},
  year={2021},
  organization={PMLR}
}

@article{stadie2018some,
  title={Some considerations on learning to explore via meta-reinforcement learning},
  author={Stadie, Bradly C and Yang, Ge and Houthooft, Rein and Chen, Xi and Duan, Yan and Wu, Yuhuai and Abbeel, Pieter and Sutskever, Ilya},
  journal={arXiv preprint arXiv:1803.01118},
  year={2018}
}

@inproceedings{zintgraf2021exploration,
  title={Exploration in approximate hyper-state space for meta reinforcement learning},
  author={Zintgraf, Luisa M and Feng, Leo and Lu, Cong and Igl, Maximilian and Hartikainen, Kristian and Hofmann, Katja and Whiteson, Shimon},
  booktitle={International Conference on Machine Learning},
  pages={12991--13001},
  year={2021},
  organization={PMLR}
}

@article{grigsby2020measuring,
  title={Measuring visual generalization in continuous control from pixels},
  author={Grigsby, Jake and Qi, Yanjun},
  journal={arXiv preprint arXiv:2010.06740},
  year={2020}
}

@article{dao2022flashattention,
  title={FlashAttention: Fast and Memory-Efficient Exact Attention with IO-Awareness},
  author={Dao, Tri and Fu, Daniel Y and Ermon, Stefano and Rudra, Atri and R{\'e}, Christopher},
  journal={arXiv preprint arXiv:2205.14135},
  year={2022}
}

@article{levine2020offline,
  title={Offline reinforcement learning: Tutorial, review, and perspectives on open problems},
  author={Levine, Sergey and Kumar, Aviral and Tucker, George and Fu, Justin},
  journal={arXiv preprint arXiv:2005.01643},
  year={2020}
}

@inproceedings{zhang2021metacure,
  title={Metacure: Meta reinforcement learning with empowerment-driven exploration},
  author={Zhang, Jin and Wang, Jianhao and Hu, Hao and Chen, Tong and Chen, Yingfeng and Fan, Changjie and Zhang, Chongjie},
  booktitle={International Conference on Machine Learning},
  pages={12600--12610},
  year={2021},
  organization={PMLR}
}

@article{beck2023survey,
  title={A Survey of Meta-Reinforcement Learning},
  author={Beck, Jacob and Vuorio, Risto and Liu, Evan Zheran and Xiong, Zheng and Zintgraf, Luisa and Finn, Chelsea and Whiteson, Shimon},
  journal={arXiv preprint arXiv:2301.08028},
  year={2023}
}

@article{duan2016rl,
  title={RL {$^2$}: Fast reinforcement learning via slow reinforcement learning},
  author={Duan, Yan and Schulman, John and Chen, Xi and Bartlett, Peter L and Sutskever, Ilya and Abbeel, Pieter},
  journal={arXiv preprint arXiv:1611.02779},
  year={2016}
}

@phdthesis{zintgraf2022fast,
  title={Fast adaptation via meta reinforcement learning},
  author={Zintgraf, Luisa},
  year={2022},
  school={University of Oxford}
}

@article{humplik2019meta,
  title={Meta reinforcement learning as task inference},
  author={Humplik, Jan and Galashov, Alexandre and Hasenclever, Leonard and Ortega, Pedro A and Teh, Yee Whye and Heess, Nicolas},
  journal={arXiv preprint arXiv:1905.06424},
  year={2019}
}

@article{wang2016learning,
  title={Learning to reinforcement learn},
  author={Wang, Jane X and Kurth-Nelson, Zeb and Tirumala, Dhruva and Soyer, Hubert and Leibo, Joel Z and Munos, Remi and Blundell, Charles and Kumaran, Dharshan and Botvinick, Matt},
  journal={arXiv preprint arXiv:1611.05763},
  year={2016}
}

@inproceedings{
sun2025unsupervised,
title={Unsupervised Zero-Shot Reinforcement Learning via Dual-Value Forward-Backward Representation},
author={Jingbo Sun and Songjun Tu and qichao Zhang and Haoran Li and Xin Liu and Yaran Chen and Ke Chen and Dongbin Zhao},
booktitle={The Thirteenth International Conference on Learning Representations},
year={2025},
url={https://openreview.net/forum?id=0QnKnt411O}
}

@article{jeen2025zero,
  title={Zero-shot reinforcement learning under partial observability},
  author={Jeen, Scott and Bewley, Tom and Cullen, Jonathan M},
  journal={arXiv preprint arXiv:2506.15446},
  year={2025}
}

@article{gao2023context,
  title={Context shift reduction for offline meta-reinforcement learning},
  author={Gao, Yunkai and Zhang, Rui and Guo, Jiaming and Wu, Fan and Yi, Qi and Peng, Shaohui and Lan, Siming and Chen, Ruizhi and Du, Zidong and Hu, Xing and others},
  journal={Advances in Neural Information Processing Systems},
  volume={36},
  pages={80024--80043},
  year={2023}
}

@article{bobrin2025zero,
  title={Zero-shot adaptation of behavioral foundation models to unseen dynamics},
  author={Bobrin, Maksim and Zisman, Ilya and Nikulin, Alexander and Kurenkov, Vladislav and Dylov, Dmitry},
  journal={arXiv preprint arXiv:2505.13150},
  year={2025}
}

@inproceedings{henaff2023study,
  title={A study of global and episodic bonuses for exploration in contextual mdps},
  author={Henaff, Mikael and Jiang, Minqi and Raileanu, Roberta},
  booktitle={International Conference on Machine Learning},
  pages={12972--12999},
  year={2023},
  organization={PMLR}
}

@article{jeen2024zerolow,
  title={Zero-shot reinforcement learning from low quality data},
  author={Jeen, Scott and Bewley, Tom and Cullen, Jonathan M},
  journal={Advances in Neural Information Processing Systems},
  volume={37},
  pages={16894--16942},
  year={2024}
}

@inproceedings{
laskin2021urlb,
title={{URLB}: Unsupervised Reinforcement Learning Benchmark},
author={Michael Laskin and Denis Yarats and Hao Liu and Kimin Lee and Albert Zhan and Kevin Lu and Catherine Cang and Lerrel Pinto and Pieter Abbeel},
booktitle={Thirty-fifth Conference on Neural Information Processing Systems Datasets and Benchmarks Track (Round 2)},
year={2021},
url={https://openreview.net/forum?id=lwrPkQP_is}
}

@article{zhou2025comprehensive,
  title={A comprehensive survey on pretrained foundation models: A history from bert to chatgpt},
  author={Zhou, Ce and Li, Qian and Li, Chen and Yu, Jun and Liu, Yixin and Wang, Guangjing and Zhang, Kai and Ji, Cheng and Yan, Qiben and He, Lifang and others},
  journal={International Journal of Machine Learning and Cybernetics},
  volume={16},
  number={12},
  pages={9851--9915},
  year={2025},
  publisher={Springer}
}

@article{nikulin2024xland,
  title={XLand-minigrid: Scalable meta-reinforcement learning environments in JAX},
  author={Nikulin, Alexander and Kurenkov, Vladislav and Zisman, Ilya and Agarkov, Artem and Sinii, Viacheslav and Kolesnikov, Sergey},
  journal={Advances in Neural Information Processing Systems},
  volume={37},
  pages={43809--43835},
  year={2024}
}

@inproceedings{pong2022offline,
  title={Offline meta-reinforcement learning with online self-supervision},
  author={Pong, Vitchyr H and Nair, Ashvin V and Smith, Laura M and Huang, Catherine and Levine, Sergey},
  booktitle={International Conference on Machine Learning},
  pages={17811--17829},
  year={2022},
  organization={Pmlr}
}

@article{grigsby2024amago,
  title={Amago-2: Breaking the multi-task barrier in meta-reinforcement learning with transformers},
  author={Grigsby, Jake and Sasek, Justin and Parajuli, Samyak and Adebi, Daniel and Zhang, Amy and Zhu, Yuke},
  journal={Advances in Neural Information Processing Systems},
  volume={37},
  pages={87473--87508},
  year={2024}
}

@inproceedings{bruce2024genie,
  title={Genie: Generative interactive environments},
  author={Bruce, Jake and Dennis, Michael D and Edwards, Ashley and Parker-Holder, Jack and Shi, Yuge and Hughes, Edward and Lai, Matthew and Mavalankar, Aditi and Steigerwald, Richie and Apps, Chris and others},
  booktitle={Forty-first International Conference on Machine Learning},
  year={2024}
}

@article{che2024gamegen,
  title={Gamegen-x: Interactive open-world game video generation},
  author={Che, Haoxuan and He, Xuanhua and Liu, Quande and Jin, Cheng and Chen, Hao},
  journal={arXiv preprint arXiv:2411.00769},
  year={2024}
}

@article{karten2026automatic,
  title={Automatic Generation of High-Performance RL Environments},
  author={Karten, Seth and Appapogu, Rahul Dev and Jin, Chi},
  journal={arXiv preprint arXiv:2603.12145},
  year={2026}
}

@article{schwarzer2020data,
  title={Data-efficient reinforcement learning with self-predictive representations},
  author={Schwarzer, Max and Anand, Ankesh and Goel, Rishab and Hjelm, R Devon and Courville, Aaron and Bachman, Philip},
  journal={arXiv preprint arXiv:2007.05929},
  year={2020}
}

@article{oord2018representation,
  title={Representation learning with contrastive predictive coding},
  author={Oord, Aaron van den and Li, Yazhe and Vinyals, Oriol},
  journal={arXiv preprint arXiv:1807.03748},
  year={2018}
}

@article{grill2020bootstrap,
  title={Bootstrap your own latent-a new approach to self-supervised learning},
  author={Grill, Jean-Bastien and Strub, Florian and Altch{\'e}, Florent and Tallec, Corentin and Richemond, Pierre and Buchatskaya, Elena and Doersch, Carl and Avila Pires, Bernardo and Guo, Zhaohan and Gheshlaghi Azar, Mohammad and others},
  journal={Advances in neural information processing systems},
  volume={33},
  pages={21271--21284},
  year={2020}
}

@article{zheng2026can,
  title={Can We Really Learn One Representation to Optimize All Rewards?},
  author={Zheng, Chongyi and Jayanth, Royina Karegoudra and Eysenbach, Benjamin},
  journal={arXiv preprint arXiv:2602.11399},
  year={2026}
}

@article{bagatella2026soft,
  title={Soft Forward-Backward Representations for Zero-shot Reinforcement Learning with General Utilities},
  author={Bagatella, Marco and Rupf, Thomas and Martius, Georg and Krause, Andreas},
  journal={arXiv preprint arXiv:2602.06769},
  year={2026}
}

@article{chua2024learning,
  title={Learning successor features the simple way},
  author={Chua, Raymond and Ghosh, Arna and Kaplanis, Christos and Richards, Blake A and Precup, Doina},
  journal={Advances in Neural Information Processing Systems},
  volume={37},
  pages={49957--50030},
  year={2024}
}

@article{badia2020never,
  title={Never give up: Learning directed exploration strategies},
  author={Badia, Adri{\`a} Puigdom{\`e}nech and Sprechmann, Pablo and Vitvitskyi, Alex and Guo, Daniel and Piot, Bilal and Kapturowski, Steven and Tieleman, Olivier and Arjovsky, Mart{\'\i}n and Pritzel, Alexander and Bolt, Andew and others},
  journal={arXiv preprint arXiv:2002.06038},
  year={2020}
}

@inproceedings{
Raileanu2020RIDE:,
title={RIDE: Rewarding Impact-Driven Exploration for Procedurally-Generated Environments},
author={Roberta Raileanu and Tim Rocktäschel},
booktitle={International Conference on Learning Representations},
year={2020},
url={https://openreview.net/forum?id=rkg-TJBFPB}
}

@inproceedings{seo2021state,
  title={State entropy maximization with random encoders for efficient exploration},
  author={Seo, Younggyo and Chen, Lili and Shin, Jinwoo and Lee, Honglak and Abbeel, Pieter and Lee, Kimin},
  booktitle={International conference on machine learning},
  pages={9443--9454},
  year={2021},
  organization={PMLR}
}

@article{henaff2022exploration,
  title={Exploration via elliptical episodic bonuses},
  author={Henaff, Mikael and Raileanu, Roberta and Jiang, Minqi and Rockt{\"a}schel, Tim},
  journal={Advances in Neural Information Processing Systems},
  volume={35},
  pages={37631--37646},
  year={2022}
}

@inproceedings{
jajoo2026regularized,
title={Regularized Latent Dynamics Prediction is a Strong Baseline For Behavioral Foundation Models},
author={Pranaya Jajoo and Harshit Sikchi and Siddhant Agarwal and Amy Zhang and Scott Niekum and Martha White},
booktitle={The Fourteenth International Conference on Learning Representations},
year={2026},
url={https://openreview.net/forum?id=jdL6WB5jHZ}
}

@inproceedings{agarwal2025proto,
  title={Proto Successor Measure: Representing the Behavior Space of an RL Agent},
  author={Agarwal, Siddhant and Sikchi, Harshit and Stone, Peter and Zhang, Amy},
  booktitle={International Conference on Machine Learning},
  pages={566--586},
  year={2025},
  organization={PMLR}
}

@article{li2025bfm,
  title={Bfm-zero: A promptable behavioral foundation model for humanoid control using unsupervised reinforcement learning},
  author={Li, Yitang and Luo, Zhengyi and Zhang, Tonghe and Dai, Cunxi and Kanervisto, Anssi and Tirinzoni, Andrea and Weng, Haoyang and Kitani, Kris and Guzek, Mateusz and Touati, Ahmed and others},
  journal={arXiv preprint arXiv:2511.04131},
  year={2025}
}

@inproceedings{tirinzonizero,
  title={Zero-Shot Whole-Body Humanoid Control via Behavioral Foundation Models},
  author={Tirinzoni, Andrea and Touati, Ahmed and Farebrother, Jesse and Guzek, Mateusz and Kanervisto, Anssi and Xu, Yingchen and Lazaric, Alessandro and Pirotta, Matteo},
  booktitle={The Thirteenth International Conference on Learning Representations},
  year={2025}
}

@article{mittal2025isaac,
  title={Isaac lab: A gpu-accelerated simulation framework for multi-modal robot learning},
  author={Mittal, Mayank and Roth, Pascal and Tigue, James and Richard, Antoine and Zhang, Octi and Du, Peter and Serrano-Munoz, Antonio and Yao, Xinjie and Zurbr{\"u}gg, Ren{\'e} and Rudin, Nikita and others},
  journal={arXiv preprint arXiv:2511.04831},
  year={2025}
}

@article{makoviychuk2021isaac,
  title={Isaac gym: High performance gpu-based physics simulation for robot learning},
  author={Makoviychuk, Viktor and Wawrzyniak, Lukasz and Guo, Yunrong and Lu, Michelle and Storey, Kier and Macklin, Miles and Hoeller, David and Rudin, Nikita and Allshire, Arthur and Handa, Ankur and others},
  journal={arXiv preprint arXiv:2108.10470},
  year={2021}
}

@article{joshi2025benchmarking,
  title={Benchmarking massively parallelized multi-task reinforcement learning for robotics tasks},
  author={Joshi, Viraj and Xu, Zifan and Liu, Bo and Stone, Peter and Zhang, Amy},
  journal={arXiv preprint arXiv:2507.23172},
  year={2025}
}

@article{mclean2025meta,
  title={Meta-world+: An improved, standardized, rl benchmark},
  author={McLean, Reginald and Chatzaroulas, Evangelos and McCutcheon, Luc and R{\"o}der, Frank and Yu, Tianhe and He, Zhanpeng and Zentner, KR and Julian, Ryan and Terry, Jordan K and Woungang, Isaac and others},
  journal={arXiv preprint arXiv:2505.11289},
  year={2025}
}

@article{su2024roformer,
  title={Roformer: Enhanced transformer with rotary position embedding},
  author={Su, Jianlin and Ahmed, Murtadha and Lu, Yu and Pan, Shengfeng and Bo, Wen and Liu, Yunfeng},
  journal={Neurocomputing},
  volume={568},
  pages={127063},
  year={2024},
  publisher={Elsevier}
}

@article{li2024towards,
  title={Towards an information theoretic framework of context-based offline meta-reinforcement learning},
  author={Li, Lanqing and Zhang, Hai and Zhang, Xinyu and Zhu, Shatong and Yu, Yang and Zhao, Junqiao and Heng, Pheng-Ann},
  journal={Advances in Neural Information Processing Systems},
  volume={37},
  pages={75642--75667},
  year={2024}
}

@article{rupf2025optimistic,
  title={Optimistic task inference for behavior foundation models},
  author={Rupf, Thomas and Bagatella, Marco and Vlastelica, Marin and Krause, Andreas},
  journal={arXiv preprint arXiv:2510.20264},
  year={2025}
}

@article{touati2021learning,
  title={Learning one representation to optimize all rewards},
  author={Touati, Ahmed and Ollivier, Yann},
  journal={Advances in Neural Information Processing Systems},
  volume={34},
  pages={13--23},
  year={2021}
}

@article{touati2022does,
  title={Does zero-shot reinforcement learning exist?},
  author={Touati, Ahmed and Rapin, J{\'e}r{\'e}my and Ollivier, Yann},
  journal={arXiv preprint arXiv:2209.14935},
  year={2022}
}

@inproceedings{
agarwal2025a,
title={A Unified Framework for Unsupervised Reinforcement Learning Algorithms},
author={Siddhant Agarwal and Caleb Chuck and Harshit Sikchi and Jiaheng Hu and Max Rudolph and Scott Niekum and Peter Stone and Amy Zhang},
booktitle={Workshop on Reinforcement Learning Beyond Rewards @ Reinforcement Learning Conference 2025},
year={2025},
url={https://openreview.net/forum?id=tQtO75p5HA}
}

@article{gupta2018unsupervised,
  title={Unsupervised meta-learning for reinforcement learning},
  author={Gupta, Abhishek and Eysenbach, Benjamin and Finn, Chelsea and Levine, Sergey},
  journal={arXiv preprint arXiv:1806.04640},
  year={2018}
}

@article{di2025unified,
  title={A Unified Framework for Zero-Shot Reinforcement Learning},
  author={Di Ventura, Jacopo and Kleuker, Jan Felix and Plaat, Aske and Moerland, Thomas},
  journal={arXiv preprint arXiv:2510.20542},
  year={2025}
}

@article{bagatella2025td,
  title={Td-jepa: Latent-predictive representations for zero-shot reinforcement learning},
  author={Bagatella, Marco and Pirotta, Matteo and Touati, Ahmed and Lazaric, Alessandro and Tirinzoni, Andrea},
  journal={arXiv preprint arXiv:2510.00739},
  year={2025}
}

@article{park2024foundation,
  title={Foundation policies with hilbert representations},
  author={Park, Seohong and Kreiman, Tobias and Levine, Sergey},
  journal={arXiv preprint arXiv:2402.15567},
  year={2024}
}

@article{harvey2020robust,
author    = {Félix G. Harvey and Mike Yurick and Derek Nowrouzezahrai and Christopher Pal},
title     = {Robust Motion In-Betweening},
journal   = {ACM Transactions on Graphics (Proceedings of ACM SIGGRAPH)},
publisher = {ACM},
volume    = {39},
number    = {4},
year      = {2020}
}

@article{jabri2019unsupervised,
  title={Unsupervised curricula for visual meta-reinforcement learning},
  author={Jabri, Allan and Hsu, Kyle and Gupta, Abhishek and Eysenbach, Ben and Levine, Sergey and Finn, Chelsea},
  journal={Advances in Neural Information Processing Systems},
  volume={32},
  year={2019}
}

@article{pappalardo2026unsupervised,
  title={Unsupervised Learning of Efficient Exploration: Pre-training Adaptive Policies via Self-Imposed Goals},
  author={Pappalardo, Octavio},
  journal={arXiv preprint arXiv:2601.19810},
  year={2026}
}

@inproceedings{booth2023perils,
  title={The perils of trial-and-error reward design: misdesign through overfitting and invalid task specifications},
  author={Booth, Serena and Knox, W Bradley and Shah, Julie and Niekum, Scott and Stone, Peter and Allievi, Alessandro},
  booktitle={Proceedings of the AAAI Conference on Artificial Intelligence},
  volume={37},
  number={5},
  pages={5920--5929},
  year={2023}
}

@article{knox2023reward,
  title={Reward (mis) design for autonomous driving},
  author={Knox, W Bradley and Allievi, Alessandro and Banzhaf, Holger and Schmitt, Felix and Stone, Peter},
  journal={Artificial Intelligence},
  volume={316},
  pages={103829},
  year={2023},
  publisher={Elsevier}
}

@inproceedings{karwowski2023goodhart,
  title={Goodhart's law in reinforcement learning},
  author={Karwowski, Jacek and Hayman, Oliver and Bai, Xingjian and Kiendlhofer, Klaus and Griffin, Charlie and Skalse, Joar Max Viktor},
  booktitle={The Twelfth International Conference on Learning Representations},
  year={2023}
}

@article{yarats2022exorl,
  title={Don't Change the Algorithm, Change the Data: Exploratory Data for Offline Reinforcement Learning},
  author={Yarats, Denis and Brandfonbrener, David and Liu, Hao and Laskin, Michael and Abbeel, Pieter and Lazaric, Alessandro and Pinto, Lerrel},
  journal={arXiv preprint arXiv:2201.13425},
  year={2022}
}

@book{puterman2014markov,
  title={Markov decision processes: discrete stochastic dynamic programming},
  author={Puterman, Martin L},
  year={2014},
  publisher={John Wiley \& Sons}
}

@inproceedings{luo2023perpetual,
  title={Perpetual humanoid control for real-time simulated avatars},
  author={Luo, Zhengyi and Cao, Jinkun and Kitani, Kris and Xu, Weipeng and others},
  booktitle={Proceedings of the IEEE/CVF International Conference on Computer Vision},
  pages={10895--10904},
  year={2023}
}

\newpage
\startcontents[appendix]

\appendix

\printcontents[appendix]{}{1}{\section*{Appendix Table of Contents}\setcounter{tocdepth}{2}}

\newpage

\section*{Appendix}

\section{Additional Background}
\label{app:background}

\subsection{Reinforcement Learning}
\label{app:background:rl}

A Markov Decision Process (MDP) \citep{puterman2014markov} is defined as a tuple $(\mathcal{S}, \mathcal{A}, \mu, P, r)$ where $\mathcal{S}$ is called the state space, $\mathcal{A}$ is the action space, $\mu$ is the initial state distribution, $P:\mathcal{S} \times \mathcal{A} \rightarrow \Delta(\mathcal{S})$ is the transition function that maps a given state and action to a distribution of next states and $r:\mathcal{S}\rightarrow \mathbb{R}$ is the scalar reward function. The agent executes a policy, $\pi: \mathcal{S}\rightarrow \Delta(\mathcal{A})$, to interact with the environment. The Reinforcement Learning (RL) objective is to find the optimal policy, $\pi^*$, defined as, 

\begin{equation}
    \pi^* = \text{argmax}_\pi \mathbb{E}_\pi \left[ \sum_t \gamma^t r(s_t) \right]
\end{equation}

\subsection{Behavior Foundation Models}
\label{app:background:zsrl}

A Behavior Foundation Model (BFM), for a given dynamical system, allows an agent to produce near-optimal policies for any given reward function using little or no additional computation. As a result, this paradigm has often been referred to as \emph{Zero-Shot Reinforcement Learning}. Specifically, in this work, we consider successor-measure-based BFMs that learn to represent successor measures for a wide range of policies. Mathematically, the successor measure for a policy is defined as, 

\begin{equation}
    M^\pi(s, a, X) = \mathbb{E}_{\pi} \left[ \sum_{t=0}^\infty \gamma^t \mathbb{P}^\pi (s_t \in X | s_0 = s, a_0 = a) \right], \forall X \subset \mathcal{S}.
\end{equation}

BFMs represent $M^{\pi_z}$ (successor measure for policy parameterized using $z$) as $M^{\pi_z}(s, a, s^+) = \psi(s, a, z)^T \phi(s^+)$. Connections have been drawn \citep{touati2021learning, agarwal2025proto, agarwal2025a} between this representation and successor features with $\psi$ being the successor feature for the state feature $\mathbb{E}_\rho[\phi \phi^T] \phi(s)$. Methods like \cite{touati2021learning, agarwal2025proto, jajoo2026regularized}  use the following objective to learn these representations, 

\begin{multline}
\label{eq:fb_loss}
\mathcal{L}(\phi, \psi) = -\mathbb{E}_{s, a, s' \sim \rho}[\psi(s, a, z)^T \phi(s')] 
    \\+ \frac{1}{2} \mathbb{E}_{s, a, s' \sim \rho
    ,s^+ \sim \rho}[(\psi(s, a, z)^T \phi(s^+) - \gamma\bar \psi(s', \pi_z(s'), z)^T \bar \phi(s^+))^2]
\end{multline}

where $\rho$ is the offline dataset. These BFMs consider tasks defined by rewards, $r_z(s) = \phi(s)^T z$ for which the corresponding optimal Q function can be written as, $Q^*_z(s, a) = \psi(s, a, z)^T z$ with the optimal policy $\pi_z = \text{argmax}_a Q^*_z(s, a) = \text{argmax}_a \psi(s, a, z)^T z$. Overall, BFMs can be represented using $(\phi, \psi, \pi_z)$. All three networks $\phi$, $\psi$ and $\pi_z$ are learned using reward-free interactions with $\phi$ and $\psi$ being trained using Equation \ref{eq:fb_loss} and $\pi_z$ as the optimal policy for the reward $r_z(s) = \phi(s)^T z$.

At inference, when a reward function is provided, the BFMs simply need to find the $z$ for the given reward function as the corresponding policy $\pi_z$ is already trained to be optimal for $r_z$. This $z$ can be inferred simply by linear regression on reward samples, i.e. $z^* = \text{argmin}_z \mathbb{E}_\rho [(r(s) - \phi(s)^T z)^2]$.
Several methods~\citep{touati2021learning, agarwal2025proto, jajoo2026regularized, park2024foundation, bagatella2025td} have been proposed with a similar construction of BFMs.

\section{Environment Details}
\label{app:additional_envs}

\subsection{ColorGrid}
ColorGrid is an $n \times n$ continuous 2D navigation task in which a hidden reward grid $c_r$ assigns a scalar payoff to each unit cell. The agent does not observe its true position; instead, it sees the cell label scrambled by a hidden permutation $c_d$. A task is the pair $(c_d, c_r)$, so the agent must resolve two independent unknowns within a trial: how its movements map to observations, and which cell pays the most.

\begin{wrapfigure}{r}{0.25\linewidth}
    \vspace{-4mm}
    \centering
    \includegraphics[width=\linewidth]{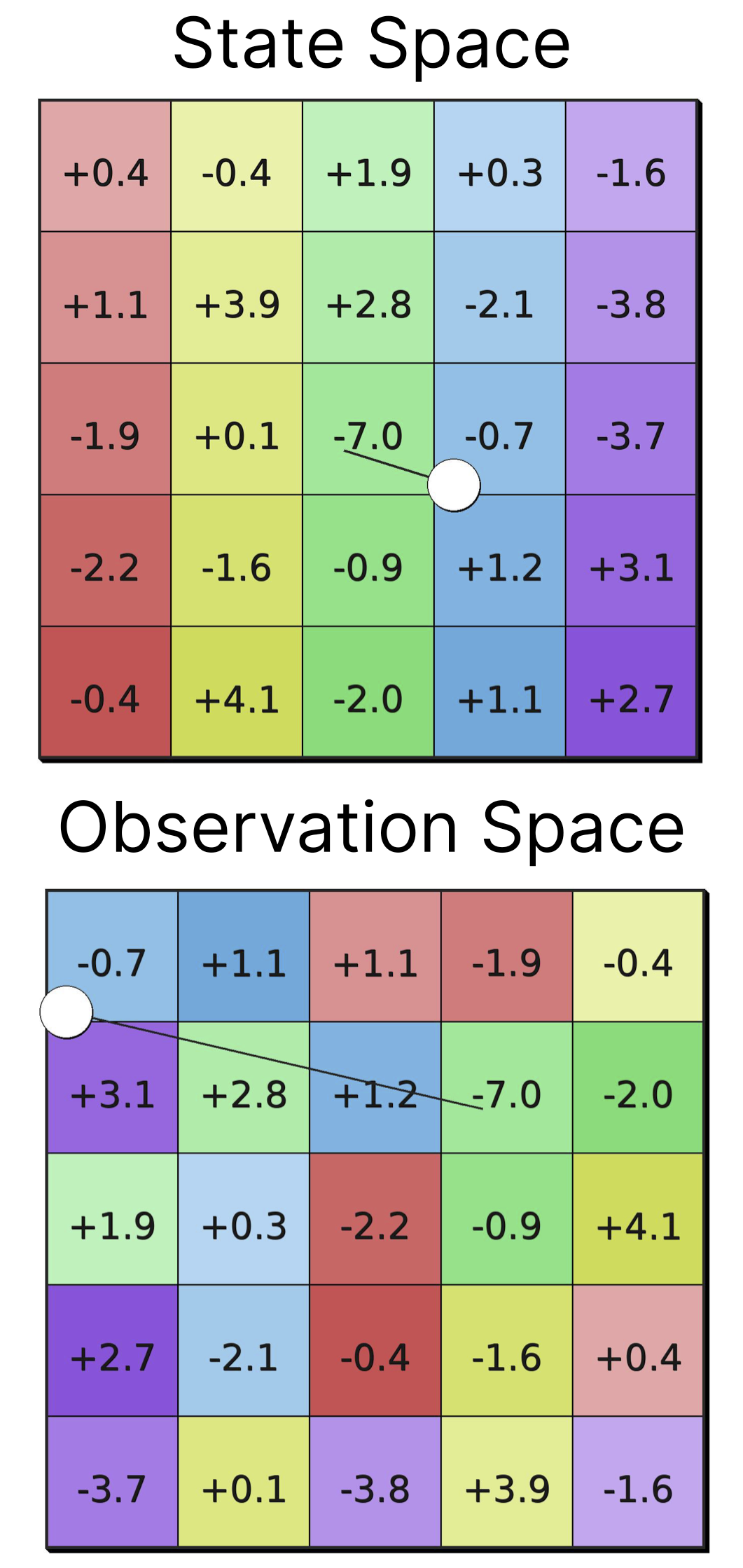}
    \vspace{-3mm}
    \caption{An example ColorGrid transition for $n=5$.}
    \label{fig:colorgrid}
    \vspace{-5mm}
\end{wrapfigure}
\paragraph{Dynamics and reward.} 
The true state is a continuous position $s_t \in [0, n]^2$. A 2D velocity action $a_t \in [-1, 1]^2$ updates it as $s_{t+1} = \mathrm{clip}(s_t + a_t, 0, n)$, and the reward is the value of the current true cell, $r_t = c_r[\lfloor s_t^{(y)} \rfloor, \lfloor s_t^{(x)} \rfloor]$. Movement and reward follow the true grid; $c_d$ only changes what the agent observes. The observation relabels the integer part of the position by $c_d$ and keeps the fractional part, so motion within a cell looks smooth, but crossing a cell boundary makes the observed position jump to an unrelated cell (Figure~\ref{fig:colorgrid}). To navigate, the agent must implicitly invert the permutation. Inner episodes last $60$ steps for $n = 5$ and $80$ steps for $n = 9$. We report oracle-normalized return: per-episode return divided by that of an oracle that teleports from the random start cell to $\arg\max c_r$ and stays there, so a score of $1.0$ matches the oracle.

\paragraph{Observations.}
The observation is three-dimensional: the scrambled 2D position described above, and a flag that is set for one step at each boundary between inner episodes of a trial. Neither $c_d$ nor $c_r$ is observed.

\paragraph{Task distribution.}
Permutations and reward grids come from two fixed, deduplicated pools, with reward entries sampled i.i.d.\ from $\mathcal{N}(0, 3)$ and clipped to $[-10, 10]$. Each pool is split into $10{,}000$ training entries and $10{,}000$ disjoint held-out entries. By default, the $i$-th task pairs the $i$-th entry of each pool, $(c_d^i, c_r^i)$. In Figure~\ref{fig:colored_grid_9x9}, tasks instead come from the cross product of the two pools, so a policy can train on many reward functions in a single permutation, or vice versa.

\subsection{Wind}

Wind is a 2D point meta-RL navigation task introduced by \citet{dorfman2020offline} and adapted by \citet{ni2022recurrent} as part of their POMDP suite. The agent commands position deltas on the plane, and the world adds a hidden, task-specific wind vector to its motion at every step. A task is an unobserved $(w, g)$ pair in which $w$ perturbs the dynamics while $g$ defines the reward landscape. The two hidden variables stress meta-RL along complementary axes. The wind is identifiable only from the difference between commanded actions and observed motion, and the goal only from reward, so a competent policy must jointly infer the dynamics and search for an unmarked target.

\paragraph{Dynamics and reward.}

The agent's state is a 2D position $s_t \in \mathbb{R}^2$, advanced by an action $a_t \in [-0.1, 0.1]^2$ and a per-task wind $w \in [-0.08, 0.08]^2$ via $s_{t+1} = s_t + a_t + w$. Inner episodes last $75$ steps and terminate early on entry into a small disk around the goal $g$. The reward is $r_t = e^{-\lVert s_t - g \rVert} + \mathbf{1}[s_t \in \text{goal disk}]$, bounded everywhere on the plane with a sharp $+1$ bonus inside the disk.

\paragraph{Observations.}
The agent receives its own 2D position together with an episode-boundary flag that fires for one step at the boundary between inner episodes within a trial. Both $w$ and $g$ are hidden.

\paragraph{Task distribution.}
During training, we sample $80$ winds uniformly from $[-0.08, 0.08]^2$ and $80$ goal locations on the unit circle, with the circle partitioned into the four standard quadrants. One quadrant ($Q_2$ in our experiments) is held out from the training goal set, so the agent is never rewarded for entering the upper-left region of the plane during training. The two test environments measure distinct generalization gaps that we report separately.

\subsection{DMC Walker}
\label{app:dmc_walker}
We use the dynamics-generalization walker from DMC Remastered \cite{grigsby2020measuring}: the DeepMind Control walker \cite{tassa2018deepmind} with its embodiment resampled from a dynamics seed. Observations are state-based, so visual randomization is disabled.

\paragraph{Dynamics and task distribution.}
Each dynamics seed defines a body by resampling torso and foot geometry, knee and ankle joint ranges, motor damping and armature, and ground friction. We draw $11{,}000$ distinct seeds once. The first $n$ form the training pool, so smaller pools nest inside larger ones, and the final $1{,}000$ form a disjoint held-out pool shared across all $n$. Section \ref{sec:metabfm} trains on $n = 10{,}000$ environments.

\paragraph{Observations and actions.}
The agent receives the 24-D proprioceptive walker state together with an episode-boundary flag that fires for one step at the boundary between inner episodes. The action is a 6-D vector of joint torques in $[-1, 1]^6$, repeated for two control steps. Inner episodes last $250$ agent steps ($500$ control steps). The body is hidden.

\paragraph{Rewards.}
Every reward lies in $[0, 1]$ per control step, and we report episode returns rescaled to the standard $0$--$1000$ DMC score. Training uses only the native \emph{walk} reward (torso height above $1.2$\,m, upright, forward speed of at least $1$\,m/s). The native \emph{stand} and \emph{run} rewards drop the speed term or raise the target to $8$\,m/s. \emph{Crouch} and \emph{crouch walk} replace the standing height condition of \emph{stand} and \emph{walk} with a two-sided band of $[0.6, 0.9]$\,m and a margin of $0.2$\,m. \emph{Walk backward} multiplies the standing height term by a velocity ramp that pays $0$ when walking forward at $1$\,m/s, $0.5$ at rest, and $1$ when walking backward at $1$\,m/s.

\subsection{MetaWorld}
\label{app:metaworld}
We base our experiments on the MetaWorld manipulation suite \cite{yu2020meta}, with the reward and goal-masking corrections of \citet{mclean2025meta}, run through the GPU-accelerated MTBench reimplementation \cite{joshi2025benchmarking} for batched simulation. Tasks are simulated in NVIDIA IsaacGym \cite{makoviychuk2021isaac} with a Franka Panda arm, and episodes are 250 steps.

\paragraph{Tasks and goals.}
MetaWorld consists of 50 single-arm tabletop manipulation task families (sliding, picking, button-pressing, door opening, etc.), each parameterized by a goal vector specifying a target site or object position drawn from a task-specific range. Each training task family exposes 100 sampled goal variations, giving ML45 a total of $4{,}500$ train configurations across 45 families and 5 held-out test families. The goal vector is masked to zero in the agent's observation at every timestep, so the agent must infer both the task family and the specific goal from interaction alone.

\paragraph{Action and observation spaces.}
The action is a 4-D continuous vector $a \in [-1, 1]^4$. Three components specify a Cartesian end-effector delta and the fourth controls the parallel gripper. The observation is a 39-D continuous vector containing the end-effector position and gripper, the position and quaternion of up to two task-relevant objects (zero-padded for tasks with fewer), the previous-timestep object pose, and the goal slot (masked as described above).

\paragraph{Reward.}
At each timestep, the agent receives the sum of a dense shaping term and a sparse success bonus. MetaWorld reward functions have changed several times: MetaWorld v2 placed rewards on a more consistent scale across tasks to avoid indirectly leaking task identity, while MTBench scaled the dense reward by $0.01$ to reduce returns and improve multi-task RL. Our experiments avoid this issue with categorical value function networks, so we restore the original dense reward scale, as dense rewards are expected to improve OptiBFM online task inference.

\paragraph{Goal-reaching reward.} Because ZSRL inference fails on the five held-out ML45 tasks (Figure~\ref{fig:ml45_sigma}), we instead evaluate a novel behavior in all $50$ train and held-out environments (Figure~\ref{fig:ml45} Right) by porting the objective of MetaWorld's \emph{reach} task. Objects and dynamics are unchanged, but the task reward is replaced by a function of the distance $d$ between the gripper and a hidden target: each step yields $10\max(0, 1 - d/1.2)$, or a success bonus of $20$ when $d < 5$\,cm. Unlike MetaWorld's reach shaping, which is scaled by the initial distance to the goal, this linear ramp gives a constant gradient across the workspace and keeps success a clear step above the dense reward. The target is resampled at the start of each trial and fixed across its episodes. We only sample targets near gripper positions visited by \texttt{MetaBFM}'s exploration-mode rollouts ($\alpha=1,\beta=0$) in the same environment. This lets us verify that every target behavior is in-distribution for the exploration data the BFM is trained on, so it is a behavior ZSRL inference could learn to recover.

\subsection{Humanoid Locomotion}
\label{app:humanoid}
Following \citet{li2025bfm}, we create environments in IsaacLab \cite{mittal2025isaac} for humanoids with locomotion tasks.
The humanoids are evaluated on motion tracking with LAFAN \cite{harvey2020robust}, a diverse and challenging motion dataset.

\begin{figure}[htbp]
    \centering
    \captionsetup{font={small}}
    \includegraphics[width=0.8\linewidth]{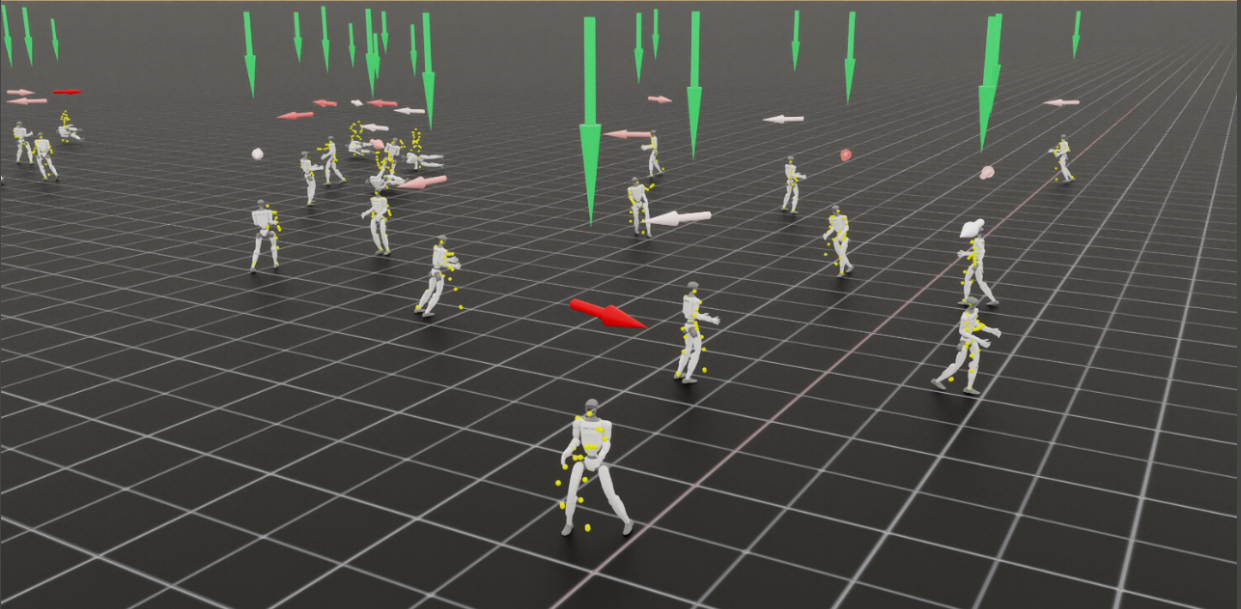}
    \caption{\textbf{Wind/Gravity illustration.} During online data collection, the wind and gravity strength in each environment are different -- the horizontal arrows indicate the direction and strength of the wind (red as the strongest, white as the weakest), while the length of vertical arrows indicate the strength of gravity.}
    \label{fig:humanoid_training}
    \vspace{-10pt}
\end{figure}

Although RL methods on robotics usually include domain randomization to achieve robust sim-to-real transfer, we intentionally try to create more dynamic and random environments, for example, different wind and gravity.
So apart from the domain randomization on mass, friction and so on, we randomly sample the scale and direction of wind, and gravity scale as shown in Figure~\ref{fig:humanoid_training}. Dynamics setting details are shown in Figure~\ref{table:domain randomization}.

\begin{figure}[h]
\centering
\begin{tabular}{@{}c@{\hspace{2em}}c@{\hspace{2em}}c@{}}
{
\footnotesize
\begin{tabular}[t]{lc}
\multicolumn{2}{c}{\textbf{Domain Randomization}} \\
\toprule
\textbf{Parameter} & \textbf{Range} \\
\midrule
COM Offset [m]            & $\mathcal{U}([-0.02, 0.02])$ \\
Link Mass             & $\mathcal{U}([0.95, 1.05])$ \\
Friction                        & $\mathcal{U}([-0.5, 1.25])$ \\
Default Joint Pos [m]    & $\mathcal{U}([-0.02, 0.02])$ \\
Push Robots [m/s]     & $\mathcal{U}([0, 0.5])$ \\
\midrule
Wind Direction[Rad]  & $\mathcal{U}([-\pi, \pi])$\\
Wind Magnitude  & $\mathcal{U}([0.0, 30.0])$\\
Gravity Magnitude[$\mathbf{g}$] & $\mathcal{U}([0.8, 1.2])$\\
\bottomrule
\end{tabular}
}
&
{
\footnotesize
\begin{tabular}[t]{lc}
\multicolumn{2}{c}{\textbf{Additive Observation Noise}} \\
\toprule
\textbf{Observation} & \textbf{Range} \\
\midrule
$q_t - \bar{q}$                & $\mathcal{U}([-0.01, 0.01])$ \\
$\dot{q}_t$                    & $\mathcal{U}([-0.5, 0.5])$ \\
$\mathrm{grav}_t$              & $\mathcal{U}([-0.05, 0.05])$ \\
$\dot{\omega}^{\mathrm{root}}_t / 4$ & $\mathcal{U}([-0.05, 0.05])$ \\
\bottomrule
\end{tabular}
}
\end{tabular}
\caption{Details in training environment.}
\label{table:domain randomization}
\end{figure}

\paragraph{State and Observation.}
The privileged state $s_t \in \mathbb{R}^{463}$ includes root height, full body pose, body orientation, as well as linear and angular velocities.
The observable state is defined as
$
o_t \triangleq \left(q_t - \bar{q}, \dot{q}_t, \tfrac{1}{4}\omega_t^{\text{root}}, g_t \right) \in \mathbb{R}^{64},
$
where $q_t \in \mathbb{R}^{29}$ is the joint position normalized by the nominal pose $\bar{q}$, $\dot{q}_t \in \mathbb{R}^{29}$ is the joint velocity, $\omega_t^{\text{root}} \in \mathbb{R}^3$ is the root angular velocity, and $g_t \in \mathbb{R}^3$ is the projected gravity.
To mitigate partial observability, we use an observable history
$
o_{t,H} \triangleq \left(o_{t-H}, a_{t-H}, \dots, o_t \right) \in \mathbb{R}^{93H + 64},
$
which concatenates past observations and actions over a horizon of length $H$.
All observation components (except root height in the privileged state) are normalized with respect to the humanoid's current facing direction and root position.
During BFM training, we assume access to a dataset of unlabeled motion trajectories $\mathcal{D} = \{\tau\}$, where each trajectory $\tau = (o_1, s_1, \dots, o_{T}, s_{T})$ contains both observable and privileged states.

\paragraph{Action.}
Following PHC~\cite{luo2023perpetual}, we use a proportional-derivative (PD) controller at each DoF of the humanoid, with the action $a_t$ specifying the PD target.
With the target joint set as $q_t^d=a_t$, the torque applied at each joint is $\tau^i=k^p \circ (a_t - q_t) - k^d \circ \dot{q}_t$.
This is different from the residual action representation used in prior motion imitation methods, where the action is added to the reference pose: $q^d_t = \hat{q}_t+a_t$ to speed up training.
 We do not use any external forces or meta-PD control.
For the 29-degree-of-freedom (DoF) humanoid we use -- Unitree G1, the action $a\in \mathbb{R}^{29}$ contains the PD controller targets for all DoFs.

\begin{figure}[htbp]
    \centering
    \captionsetup{font={small}}
    \includegraphics[width=\linewidth]{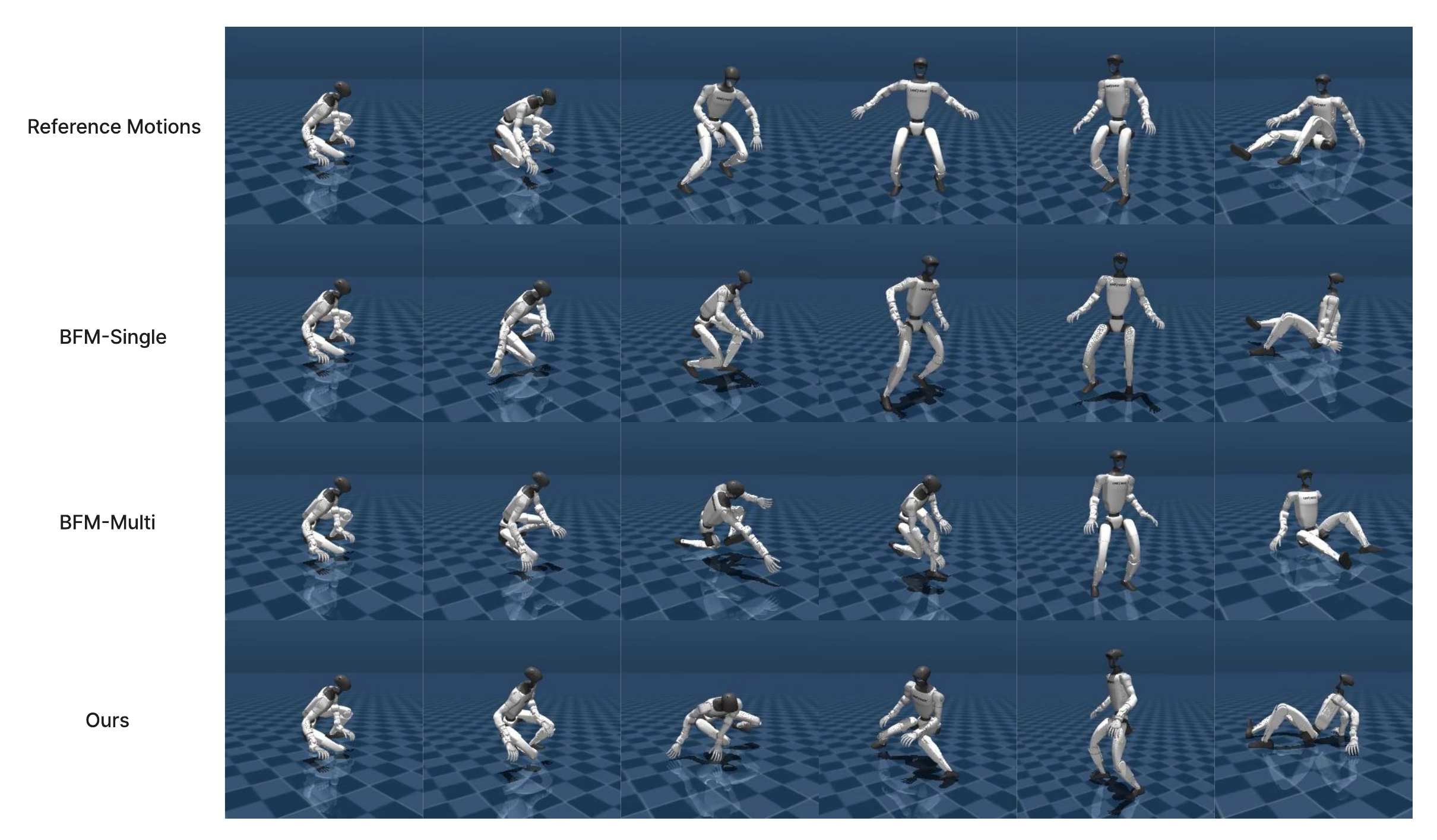}
    \caption{\textbf{Qualitative Comparison on Humanoid Motion Tracking.}}
    \label{fig:tracking_eval}
\end{figure}

\paragraph{BFM failures under multiple dynamics.}
BFMs trained with multi-dynamics data might suffer from taking actions from the wrong environment dynamics. 
This can be shown directly via the visualizations in Fig.~\ref{fig:tracking_eval}.
For BFM-Multi, in the stage of trying to get up from the ground, the robot acts as if there were wind from certain directions, loses balance and finally tracks badly. 
As a comparison, BFMs with context learning can identify the dynamics, so the states are close to the BFM-Single, which is trained solely on one type of dynamics.

\section{Implementation Details}
\label{app:implementation}

\subsection{Latent Context Learning}
\label{app:implementation:lcl}

\algrenewcommand\algorithmicindent{0pt}

\begin{algorithm}[h!]
\caption{Latent Context Learning}
\label{alg:lcl}
\textbf{Input:} trajectory batch $\tau = \{\tau^{(i)}\}_{i=1}^{B}$ with $\tau^{(i)} = \{(s_0, a_0, \dots, s_T)\}$; EMA rate $\rho$; networks $E_\theta, S_\theta, \mathrm{LDM}_\theta, \mathrm{IDM}_\theta, \mathrm{FDM}_\theta$; EMA target encoder $E^-$. \\
\textbf{Output:} augmented trajectory $\bar{\tau} = \{(\bar{s}_t, a_t, \bar{s}_{t+1})\}_t$ with $\bar{s}_t = (s_t, h_t, \tilde{c}_t)$.
\begin{algorithmic}
  \State \textit{// Augment each $\tau$ with dynamics features}
  \State $h_t = E_\theta(s_t,a_{t-2}),\;\; h^-_t = E^-(s_t,a_{t-2}),\;\; \tilde{c}_t = S_\theta(\tau_{:t}),\;\; \bar{s}_t = (s_t, h_t, \tilde{c}_t).$
  \State \textit{// Self-supervised dynamics losses}
  \State $\widehat{\mathcal{L}}_{\text{LDM}} = -\tfrac{1}{N} \sum_{i,t} \cos\!\bigl(\mathrm{LDM}_\theta(h_t, a_t, \tilde{c}_t),\; h^-_{t+1}\bigr)$
  \State $\widehat{\mathcal{L}}_{\text{IDM}} = \tfrac{1}{2N} \sum_{i,t} \bigl\| \mathrm{IDM}_\theta(h_t, h_{t+1}, \tilde{c}_t, \tilde{c}_{t+1}) - a_t \bigr\|^2$
  \State $\widehat{\mathcal{L}}_{\text{FDM}} = \tfrac{1}{2N} \sum_{i,t} \bigl\| \mathrm{FDM}_\theta(s_t, a_t, \mathrm{sg}(\tilde{c}_t)) - s_{t+1} \bigr\|^2$ \quad \textit{// $\tilde{c}_t$ detached by default (diagnostic only)}
  \State $\widehat{\mathcal{L}}_{\text{REG}} = -\tfrac{1}{B(B-1)\,T} \sum_{i \neq j,\, t} \bigl[\, 1 - \cos\!\bigl(\tilde{c}_t^{(i)},\; \tilde{c}_t^{(j)}\bigr) \,\bigr]$
  \State Take a gradient step on $\theta$ to minimize a weighted sum of (normalized) $\widehat{\mathcal{L}}_{\text{LDM}}, \widehat{\mathcal{L}}_{\text{IDM}}, \widehat{\mathcal{L}}_{\text{FDM}}, \widehat{\mathcal{L}}_{\text{REG}}$.
  \State EMA target update: $E^- \leftarrow (1-\rho)\, E^- + \rho\, E_\theta$.
  \State \textbf{return} augmented trajectory $\bar{\tau}$.
\end{algorithmic}
\end{algorithm}

Algorithm \ref{alg:lcl} summarizes the Latent Context Learning (LCL) procedure, which augments states with features used for BFM training and intrinsic rewards. Per-domain choices are summarized in Table~\ref{tab:lcl_hparams}. All four domains use small Transformer dynamics encoders. The optional cross-trajectory regularizer is disabled in Section \ref{sec:metabfm}, though preliminary experiments in Figure \ref{fig:lcl_offline} enabled it.

\begin{table}[t]
\centering
\caption{Latent Context Learning hyperparameters.}
\label{tab:lcl_hparams}
\renewcommand{\arraystretch}{1.2}
\setlength{\tabcolsep}{2pt}
\footnotesize
\begin{tabular*}{\linewidth}{@{\extracolsep{\fill}}lcccc}
\toprule
\textbf{Hyperparameter} & \textbf{ColorGrid} & \textbf{Wind} & \textbf{MetaWorld} & \textbf{DMC Walker} \\
\midrule
Latent feature dim ($\dim h$)               & 32  & 4   & 16  & 8 \\
Inverse-dynamics feature dim                & 32  & 4   & 16  & 12 \\
\addlinespace[2pt]
Inverse-dynamics MLP hidden width           & 400 & 128 & 300 & 256 \\
Latent-dynamics predictor hidden width      & 512 & 128 & 300 & 256 \\
Diagnostic probe hidden width               & 400 & 128 & 256 & 256 \\
\addlinespace[2pt]
$S_\theta$ ($d_{\text{model}}$, layers, heads)
  & Tformer (512, 6, 8)
  & Tformer (128, 2, 4)
  & Tformer (128, 2, 4) & Tformer (256, 3, 4) \\
\bottomrule
\end{tabular*}
\par\smallskip
\begin{minipage}{\linewidth}
\footnotesize\itshape Shared across domains: latent-dynamics and inverse-dynamics loss weights both 1; EMA target rate $\rho{=}5{\times}10^{-3}$; sphere-normalized features; cross-trajectory regularizer weight 0; probe gradients detached.
\end{minipage}
\end{table}

\subsection{BFM}
\label{app:implementation:bfm}

While many BFM updates could in principle be used interchangeably, our results use TD-JEPA \cite{bagatella2025td}, primarily because it shares the most surface area with off-policy actor--critic RL. The two heads $\phi$ (state representation) and $\psi$ (task representation) each use a two-tower predictor, sharing a small input trunk before splitting into per-head towers. We additionally apply a cross-trajectory orthogonality regularizer on the predicted features. A scalar reward multiplier (ten for ColorGrid and Wind, one for MetaWorld and DMC Walker) is applied to the per-step reward inside both the SF prediction target and the BLR likelihood at deployment, so it sets the natural magnitude of the regression target $r \approx \langle \psi(s'), z\rangle$ and therefore couples directly to the Thompson-sampling noise level $\sigma$ discussed in Appendix~\ref{app:implementation:online_task_inf}.

\begin{table}[t]
\centering
\caption{BFM (TD-JEPA) hyperparameters.}
\label{tab:bfm_hparams}
\renewcommand{\arraystretch}{1.2}
\setlength{\tabcolsep}{8pt}
\begin{tabular*}{\linewidth}{@{\extracolsep{\fill}}lcccc}
\toprule
\textbf{Hyperparameter} & \textbf{ColorGrid} & \textbf{Wind} & \textbf{MetaWorld} & \textbf{DMC Walker} \\
\midrule
$\phi$-encoder output dim ($\dim \phi$)      & 200  & 160 & 220  & 256 \\
$\psi$-encoder output dim ($\dim \psi$)      & 50   & 40  & 80   & 50 \\
\addlinespace[2pt]
$\phi$-predictor hidden width                & 1024 & 800 & 1024 & 1024 \\
$\psi$-predictor hidden width                & 1024 & 800 & 1024 & 1024 \\
\addlinespace[2pt]
Successor-feature discount ($\gamma_{\text{sf}}$) & 0.97 & 0.97 & 0.98 & 0.98 \\
Reward multiplier (SF target / BLR)          & 10   & 10  & 1    & 1 \\
$\phi$--$\psi$ orthogonality weight          & 0.10 & 0.25 & 0.10 & 0.10 \\
\bottomrule
\end{tabular*}
\par\smallskip
\begin{minipage}{\linewidth}
\footnotesize\itshape In reference to notation in \citet{bagatella2025td}. Shared across domains: two-tower predictor with three layers (one shared trunk plus two per-head tower layers).
\end{minipage}
\end{table}

\subsection{Online Task Inference}
\label{app:implementation:online_task_inf}

Following OptiBFM \cite{rupf2025optimistic}, we estimate the latent task vector $z^*$ online with Bayesian linear regression on the forward features $\phi$, resampling at every timestep. Our experiments use Thompson sampling rather than the optimistic alternative because it scales to the long meta-rollouts in our evaluations, and preliminary experiments found little performance gap. The posterior is governed by an observation-noise scale $\sigma$ and prior precision $\lambda$; we found $\sigma$ to be the key hyperparameter, scaling roughly linearly with the typical magnitude of the scaled per-step reward.

\begin{table}[t]
\centering
\caption{Online task inference (Thompson Sampling) hyperparameters.}
\label{tab:online_task_inf_hparams}
\renewcommand{\arraystretch}{1.2}
\setlength{\tabcolsep}{8pt}
\begin{tabular*}{\linewidth}{@{\extracolsep{\fill}}lcccc}
\toprule
\textbf{Hyperparameter} & \textbf{ColorGrid} & \textbf{Wind} & \textbf{MetaWorld} & \textbf{DMC Walker} \\
\midrule
$\sigma$ & 10 & 15 & 5 & 0.03 \\
\bottomrule
\end{tabular*}
\par\smallskip
\begin{minipage}{\linewidth}
\footnotesize\itshape Shared across domains: Bayesian linear regression with prior precision $\lambda{=}1$, posterior decay 1, exploration bias $\kappa{=}0$, posterior resampled every environment step.
\end{minipage}
\end{table}

\subsection{Intrinsic Rewards}
\label{app:implementation:intrinsic}

The intrinsic reward at each timestep is a sum of two terms: an episodic novelty bonus computed in the style of NGU \cite{badia2020never} on the latent context features, and a term proportional to the previous-timestep latent-dynamics prediction error (returned by Alg. \ref{alg:lcl}). The combined scalar is normalized and exposed to the policy as an additional observation, so the actor can directly condition on the current intrinsic signal in addition to driving the $\beta$-blended critic. 

\begin{table}[t]
\centering
\caption{Intrinsic reward hyperparameters.}
\label{tab:intrinsic_hparams}
\renewcommand{\arraystretch}{1.2}
\setlength{\tabcolsep}{3pt}
\small
\begin{tabular*}{\linewidth}{@{\extracolsep{\fill}}lcccc}
\toprule
\textbf{Hyperparameter} & \textbf{ColorGrid} & \textbf{Wind} & \textbf{MetaWorld} & \textbf{DMC Walker} \\
\midrule
Episodic novelty bonus weight                  & 1.0  & 1.0  & 1.0  & 1.0 \\
Intrinsic critic reward scale                  & 1.0  & 1.0  & 1.0  & 1.0 \\
\addlinespace[2pt]
Episodic novelty: nearest neighbors $k$        & 10 & 10 & 10 & 10 \\
Episodic novelty: cluster distance threshold   & $8{\times}10^{-3}$ & $1{\times}10^{-3}$ & $3{\times}10^{-2}$ & $8{\times}10^{-3}$ \\
Episodic novelty: maximum kernel similarity    & 8.0  & 8.0  & 8.0  & 8.0 \\
Episodic novelty: kernel $\epsilon$            & $1{\times}10^{-4}$ & $1{\times}10^{-5}$ & $3{\times}10^{-4}$ & $1{\times}10^{-4}$ \\
\bottomrule
\end{tabular*}
\par\smallskip
\begin{minipage}{\linewidth}
\end{minipage}
\end{table}

\subsection{Actor--Critic Update}
\label{app:implementation:actor_critic}

\paragraph{Sequence backbone.}
The actor and the intrinsic--extrinsic critic share a transformer trajectory encoder with rotary position embeddings \cite{su2024roformer} and sliding-window FlashAttention \cite{dao2022flashattention}.

\paragraph{Actor and Critic.}
The critic, $Q_R$, is a small ensemble of two-hot regression heads over a fixed return range \cite{hafner2023mastering}, which helps absorb the large variation in TD-target scale induced by randomizing $\alpha$ and $\beta$. Both the actor and critic heads consume a Fourier encoding of the $(\alpha, \beta)$ scalars in Equation \ref{eq:alphabetabfm} as additional input features. The actor is an MLP with a standard Tanh-Gaussian output.

\begin{table*}[t]
\centering
\caption{Actor--critic hyperparameters.}
\label{tab:actor_critic_hparams}
\renewcommand{\arraystretch}{1.2}
\setlength{\tabcolsep}{3pt}
\footnotesize
\begin{tabular*}{\textwidth}{@{\extracolsep{\fill}}lcccc}
\toprule
\textbf{Hyperparameter} & \textbf{ColorGrid} & \textbf{Wind} & \textbf{MetaWorld} & \textbf{DMC Walker} \\
\midrule
Training sequence length                    & 360 & 226 & 501 & 128 \\
Actor / critic context length               & 96  & 128 & 128 & 128 \\
Policy attention window                     & 64 timesteps & full attention & 64 timesteps & full attention \\
\addlinespace[2pt]
Actor MLP hidden width                      & 512 & 300 & 400 & 400 \\
Critic MLP hidden width                     & 300 & 300 & 400 & 300 \\
Critic output bins (two-hot)                & 64  & 64  & 96  & 256 \\
Critic return range $[r_{\min}, r_{\max}]$  & $[-400, 600]$ & $[-50, 400]$ & $[-300, 4000]$ & $[-100, 500]$ \\
\addlinespace[2pt]
Target network rate ($\tau$)                & $5{\times}10^{-3}$ & $3{\times}10^{-3}$ & $5{\times}10^{-3}$ & $5{\times}10^{-3}$ \\
Number of Fourier frequencies for $(\alpha, \beta)$ encoding & 16 & 8 & 8 & 9 \\
\bottomrule
\end{tabular*}
\par\smallskip
\begin{minipage}{\textwidth}
\end{minipage}
\end{table*}

\subsection{Computational Resources and Statistical Significance}
\label{app:implementation:compute}

All models are trained on an NVIDIA RTX A5000 GPU, with training runs for the \texttt{MetaBFM} (Section~\ref{sec:metabfm}) experiments lasting 12--36 hours depending on the domain. Shaded error bars in results figures indicate the mean and standard deviation of at least three trials, except for MetaWorld experiments, which use a single training run due to computational constraints. Results indicating test-time adaptation curves (performance vs.\ $k$) average these results over the last 3 evaluation cycles during training, then over seeds.

\section{Limitations}
\label{app:limitations}

BFM and ZSRL algorithms are an active area of research, and reported gains often depend on hyperparameters, implementation details, and benchmark choice. Our experiments use TD-JEPA as a representative state-of-the-art BFM update, but broader comparisons across BFM objectives and online task inference methods remain important future work. \texttt{MetaBFM} also introduces test-time choices through $\alpha$ and $\beta$. These parameters are useful because they let a single policy interpolate between exploration, exploitation, and ZSRL-style reward inference, but they also create a model-selection problem. In our experiments, we evaluate a few interpretable settings; in deployment, one may need validation tasks or an online rule for choosing $\alpha$ and $\beta$.

\section{Additional Results}
\label{app:additional_results}

\subsection{Online ZSRL}
\label{app:additional_results:online}

BFMs are inherently proposed as offline algorithms \citep{touati2022does, agarwal2025proto}, trained on offline exploratory datasets, which can then be used to directly infer policies for diverse reward functions. Inference solves a linear regression problem (Appendix~\ref{app:background:zsrl}) using samples drawn from the same offline datasets. Prior work \citep{sun2025unsupervised} has shown that these BFMs are not effective for online training. Methods such as \cite{li2025bfm, tirinzonizero} have used expert datasets to guide the online training of these models. 

\begin{figure}[h!]
    \centering
    \includegraphics[width=\linewidth]{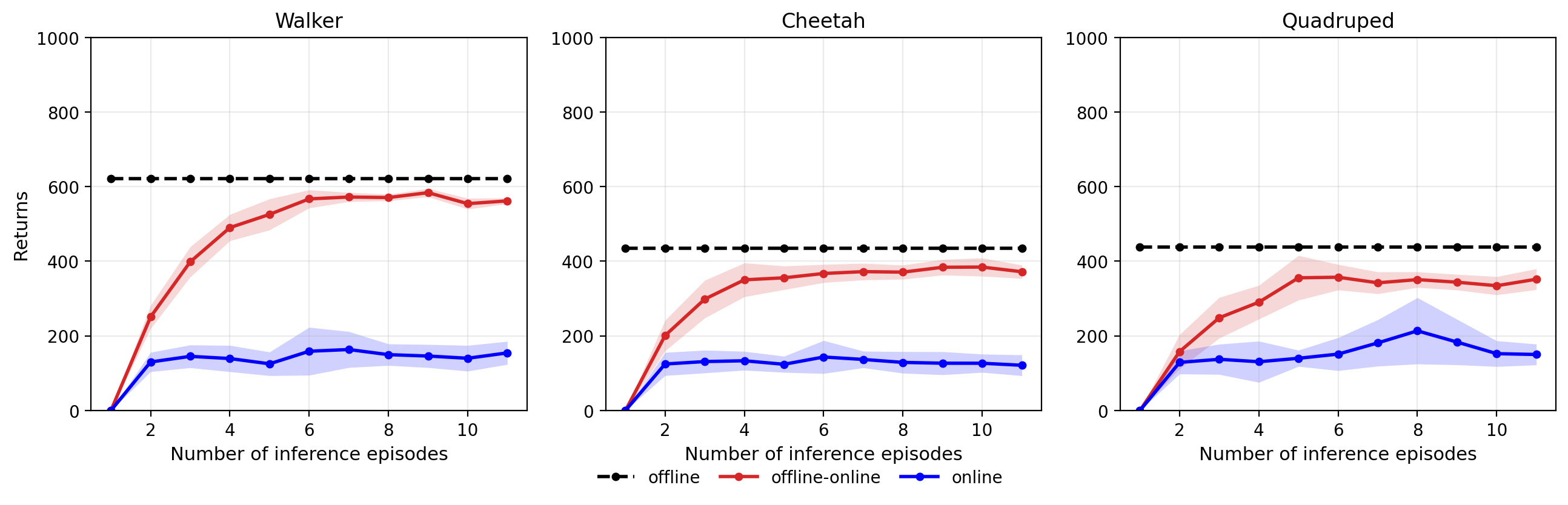}
    \caption{\textbf{Offline vs Online BFMs:} Comparison of fully offline, offline training and online inference and fully online BFMs on DM Control tasks.}
    \label{fig:dmc_online}
\end{figure}

We compare the performance of BFMs when trained offline and online on standard DM Control tasks \citep{tassa2018deepmind}. We compare (Figure \ref{fig:dmc_online}) the offline-trained BFM (on ExoRL datasets \citep{yarats2022exorl}) using offline inference (labeled \emph{offline}) with the same model using online inference (labeled \emph{offline-online}) as proposed by \cite{rupf2025optimistic} and with a fully online BFM that does not assume access to any external dataset during training or inference. 

\begin{figure}[h!]
    \centering
    \includegraphics[width=\linewidth]{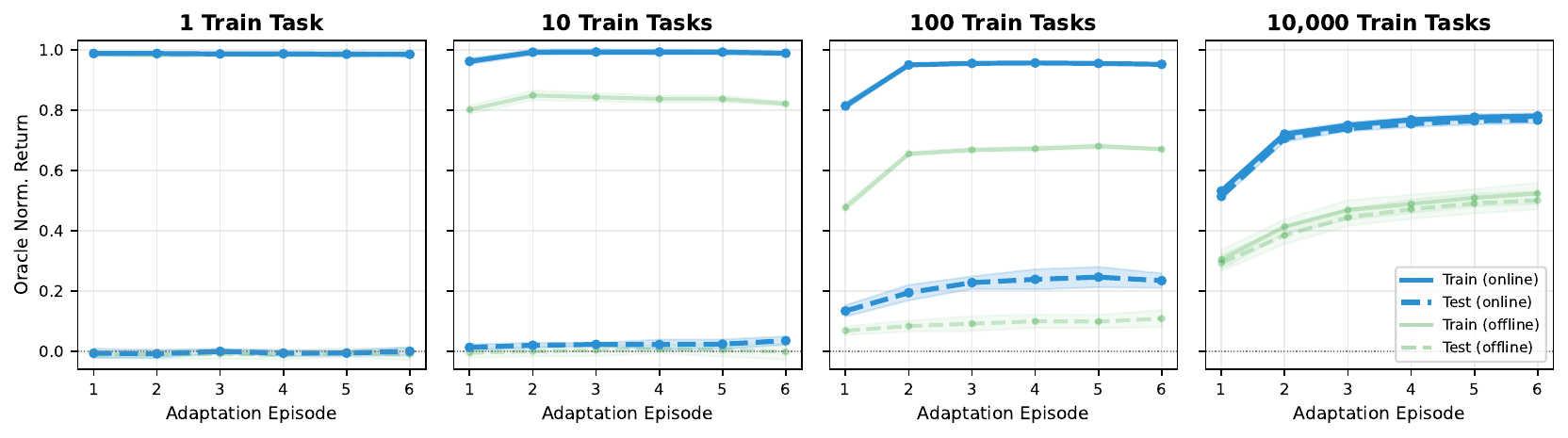}
    \caption{\textbf{Online vs. Offline Meta-RL.} In ColorGrid ($n=5$), offline meta-RL from a static dataset generated by a uniform behavior policy leads to significant distribution shift in long-context representations; online learning substantially improves performance.}
    \label{fig:app:online_vs_offline_metarl}
\end{figure}

It is clear that BFMs are optimally suited to fully offline settings. While using online inference degrades the performance slightly, the performance of online-trained BFM is significantly worse. BFMs require high coverage, which is hard to obtain during online training, leading to poor performance. On the other hand, meta-RL often improves when trained online, as observed from Figure \ref{fig:app:online_vs_offline_metarl}.

\FloatBarrier
\subsection{Utilizing Available Rewards}
\label{app:additional_results:rewards}

BFMs are designed to be trained using reward-free interactions. We design a training procedure by providing occasional reward information to the BFMs to see whether they can make use of the available rewards. On the trajectories where rewards are available, the BFMs exploit the reward function to interact with the environment using an optimal policy (optimal for the partially trained BFM). It adds regularization to the state features to ensure they remain sufficiently predictive of the available rewards, as discussed in Section \ref{sec:rews}. 

\begin{figure}[h!]
    \centering
    \includegraphics[width=0.85\linewidth]{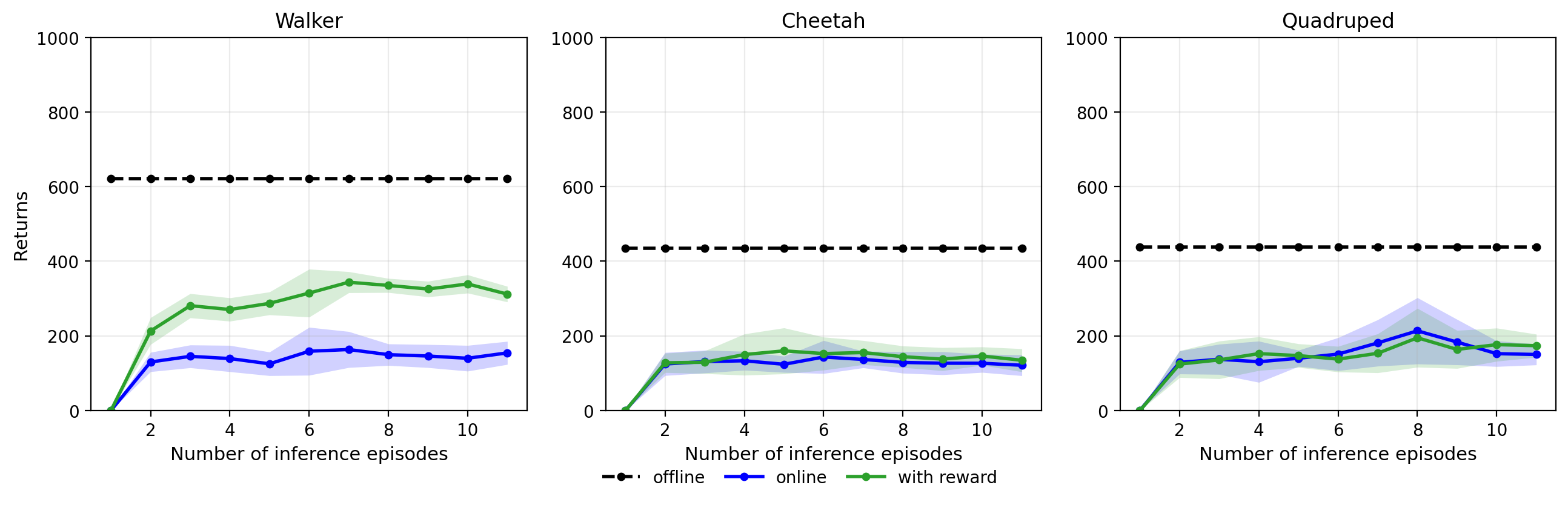}
    \caption{\textbf{Effect of using rewards during training  of BFMs:} Comparing online BFMs with and without using available rewards during training. }
    \label{fig:dmc_rews}
\end{figure}

\begin{figure}[h!]
    \centering
    \includegraphics[width=\linewidth]{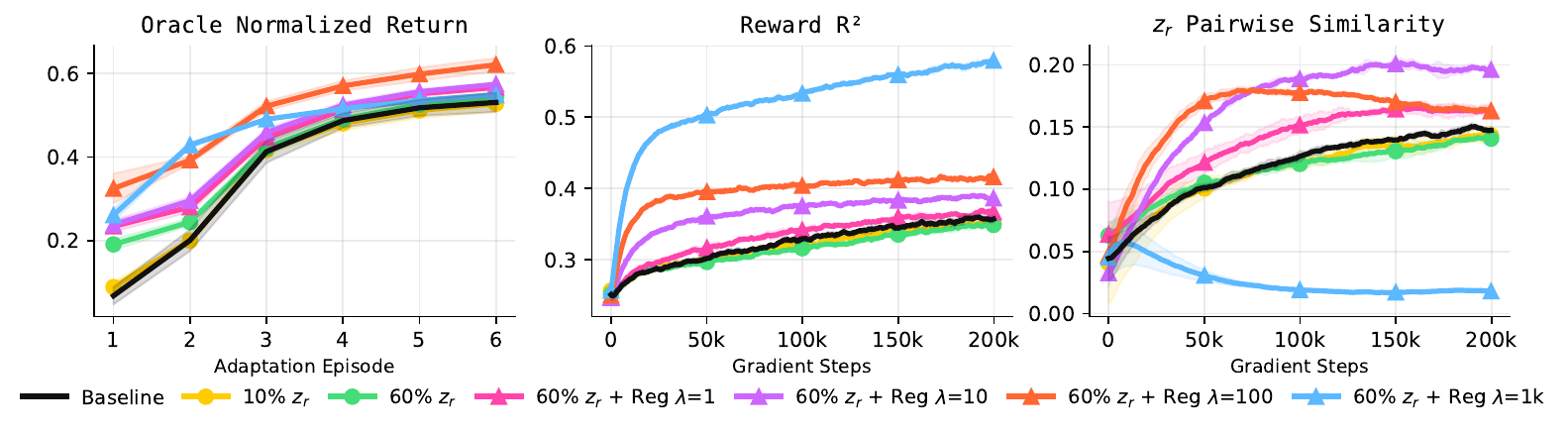}
    \caption{\textbf{Reward Regression and $z$-sampling.} Sampling reward-estimated $z$ values and fitting known reward labels (using an auxiliary loss with weight $\lambda$) fits better linear reward estimates (center), focuses learning on more similar latent tasks (right), and marginally improves early-episode adaptation in the ColorGrid domain (left).}
    \vspace{-2mm}
    \label{fig:using_rewards}
\end{figure}

We compare the performance of online-trained BFMs with and without the use of reward functions in Figure \ref{fig:dmc_rews}. The improvement in performance when using available rewards is not consistent. While an improvement is observed in the walker, similar improvements are not observed in the other two environments. On ColorGrid, the BFMs show improvement in performance when leveraging the available rewards, as shown in Figure \ref{fig:using_rewards}.

\FloatBarrier
\subsection{MetaWorld}
\label{app:additional_results:metaworld}
\begin{figure}[h!]
    \centering
    \includegraphics[width=\linewidth]{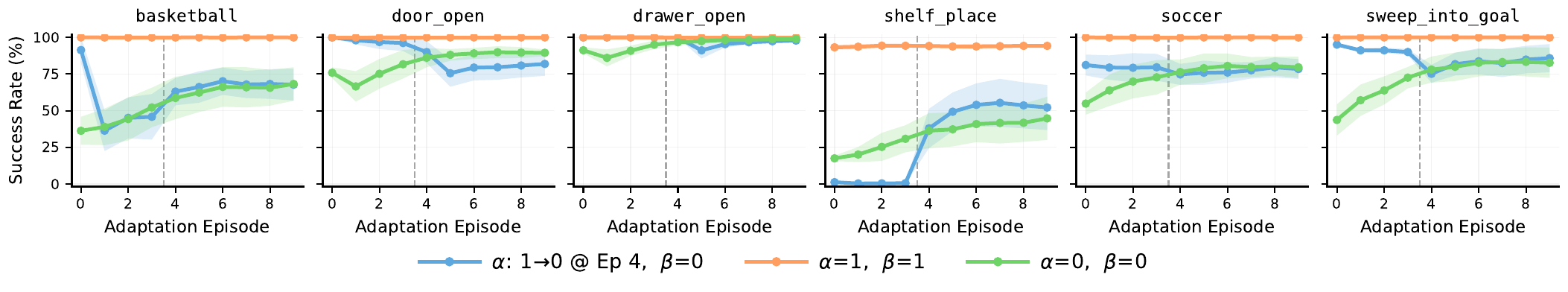}
    \caption{\textbf{Online Learning in MetaWorld ML1.} \texttt{MetaBFM} evaluated in the three main inference configurations on six MetaWorld ML1 problems.}
    \label{fig:ml1}
\end{figure}

Figure~\ref{fig:ml1} evaluates \texttt{MetaBFM} on six single-task ML1 problems, where train and test tasks share one manipulation skill and differ only in hidden goal positions.

\begin{figure}[h!]
    \centering
    \includegraphics[width=0.9\linewidth]{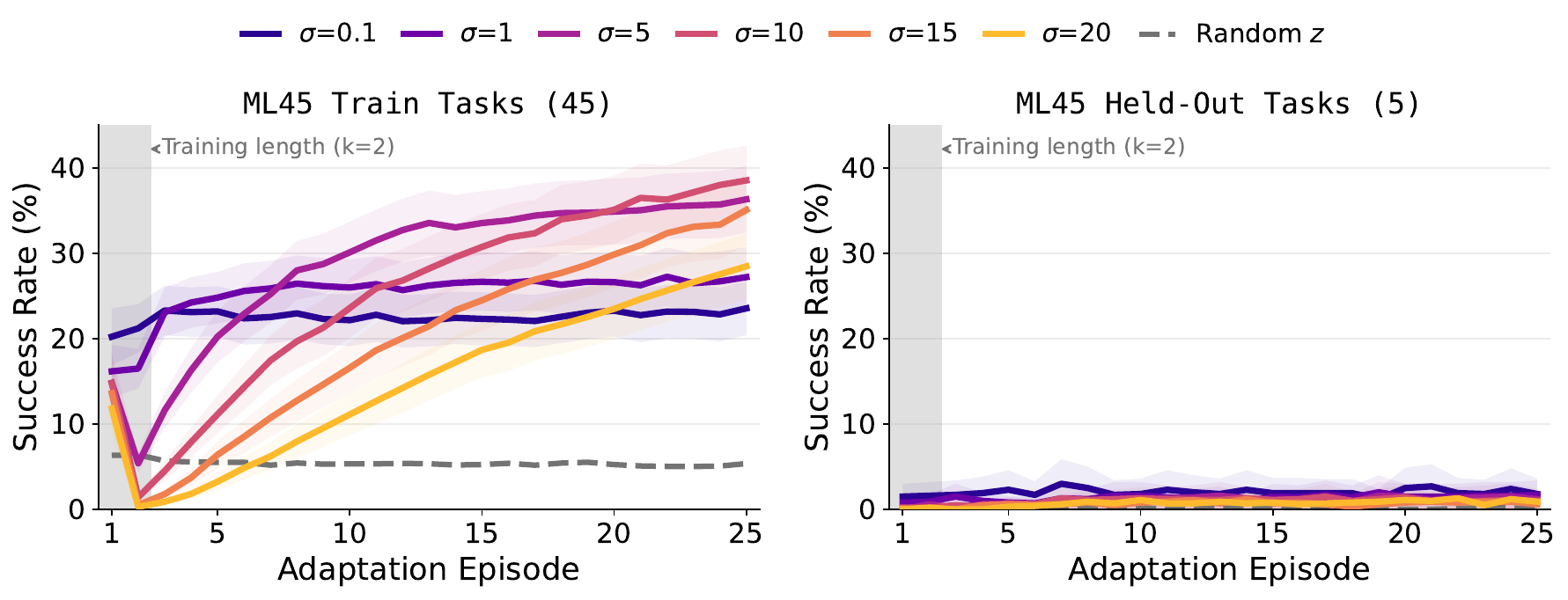}
    \caption{\textbf{ZSRL Inference Noise in ML45.} Success of pure ZSRL inference ($\alpha=0,\beta=0$) for different Thompson-sampling noise levels $\sigma$; dashed lines fix a single random $z$ for the whole trial. Larger $\sigma$ trades early success for steady improvement on the train tasks, while the held-out tasks remain unsolved.}
    \label{fig:ml45_sigma}
\end{figure}

Figure~\ref{fig:ml45_sigma} studies pure ZSRL inference ($\alpha=0,\beta=0$) on ML45 as a function of the Thompson-sampling noise $\sigma$ (Appendix~\ref{app:implementation:online_task_inf}). On the $45$ train tasks, small $\sigma$ commits to a task estimate early and plateaus, while $\sigma \in [5, 10]$ trades a worse second episode for steady improvement through $25$ episodes, well beyond the $k=2$ training horizon. On the five held-out tasks, success stays near zero at every $\sigma$.

\FloatBarrier

\end{document}